\documentclass[sigconf, nonacm, pdfa]{acmart}

\newcommand\vldbdoi{XX.XX/XXX.XX}
\newcommand\vldbpages{XXX-XXX}

\newcommand\vldbvolume{20}
\newcommand\vldbissue{3}
\newcommand\vldbyear{2026}

\newcommand\vldbauthors{\authors}
\newcommand\vldbtitle{\shorttitle} 

\newcommand\vldbavailabilityurl{https://github.com/Sean-Bin-Yang/PRISM.git}

\newcommand\vldbpagestyle{empty} 

\usepackage{multirow}
\usepackage{graphicx}
\usepackage{enumitem}
\usepackage{amsthm}
\usepackage{subcaption}
\usepackage[a-2b]{pdfx}
\newtheoremstyle{vldbdef}
{3pt}          %
{3pt}          
{\normalfont}  
{0pt}          %
{\bfseries}    
{.}            
{0.5em}       
{}

\theoremstyle{vldbdef}
\newtheorem{myDef}{Definition}

\usepackage{xcolor}

\begin{document}
\title{Rethinking Multi-View Fusion for Urban Representation Learning}

\settopmatter{authorsperrow=4}
\author{Sean Bin Yang}
\authornote{Corresponding author.}
\affiliation{%
  \institution{Aalborg University, Denmark}
}
\email{seany@cs.aau.dk}

\author{Zongyi Xu}
\affiliation{%
  \institution{Chongqing University of Posts and Telecommunications, China}
}
\email{xuzy@cqupt.edu.cn}

\author{Jilin Hu}
\affiliation{%
	\institution{East China Normal University, China}
}
\email{jlhu@dase.ecnu.edu.cn}

\author{Ying Sun}
\affiliation{%
	\institution{Chongqing University of Posts and Telecommunications, China}
}
\email{sunying@cqupt.edu.cn}

\author{Kristian Torp}
\affiliation{%
	\institution{Aalborg University, Denmark}
}
\email{torp@cs.aau.dk}
\author{Bin Yang}
\affiliation{%
  \institution{East China Normal University, China}
}
\email{byang@dase.ecnu.edu.cn}

\author{Torben Bach Pedersen}
\affiliation{%
	\institution{Aalborg University, Denmark}
}
\email{tbp@cs.aau.dk}

\author{Christian S. Jensen}
\affiliation{%
  \institution{Aalborg University, Denmark}
}

\email{csj@cs.aau.dk}

\begin{abstract}
Urban region representation learning encodes heterogeneous urban signals—such as mobility flows, points of interest, and land-use patterns—into unified embeddings for downstream tasks, including check-in prediction, crime forecasting, and service demand estimation. 
Existing approaches rely on globally shared fusion and joint optimization, implicitly assuming uniform modality importance and naturally balanced multi-objective training.
However, urban data is inherently heterogeneous. Our empirical analysis reveals two structural issues: (i) modality utility varies markedly across regions, causing static fusion to overemphasize signals from certain modalities; and (ii) heterogeneous objectives exhibit persistent gradient imbalance, causing representational bias and unstable training.
We present \texttt{PRISM}, a self-adjusting framework for adaptive multi-view learning. 
\texttt{PRISM} adapts sparse MoE routing to learn region-specific view allocations, while a PID (Proportional Integral Derivative)-inspired regulation mechanism mitigates gradient conflicts during optimization.
Experiments on three large-scale urban datasets show that \texttt{PRISM} achieves more competitive performance than strong multi-view baselines, while providing more stable optimization behavior and more balanced region representations.
\end{abstract}

\maketitle

\pagestyle{\vldbpagestyle}
\begingroup\small\noindent\raggedright\textbf{PVLDB Reference Format:}\\
\vldbauthors. \vldbtitle. PVLDB, \vldbvolume(\vldbissue): \vldbpages, \vldbyear.\\
\href{https://doi.org/\vldbdoi}{doi:\vldbdoi}
\endgroup
\begingroup
\renewcommand\thefootnote{}\footnote{\noindent
This work is licensed under the Creative Commons BY-NC-ND 4.0 International License. Visit \url{https://creativecommons.org/licenses/by-nc-nd/4.0/} to view a copy of this license. For any use beyond those covered by this license, obtain permission by emailing \href{mailto:info@vldb.org}{info@vldb.org}. Copyright is held by the owner/author(s). Publication rights licensed to the VLDB Endowment. \\
\raggedright Proceedings of the VLDB Endowment, Vol. \vldbvolume, No. \vldbissue\ %
ISSN 2150-8097. \\
\href{https://doi.org/\vldbdoi}{doi:\vldbdoi} \\
}\addtocounter{footnote}{-1}\endgroup

\ifdefempty{\vldbavailabilityurl}{}{
\vspace{.3cm}
\begingroup\small\noindent\raggedright\textbf{PVLDB Artifact Availability:}\\
The source code, data, and/or other artifacts have been made available at \url{https://github.com/Sean-Bin-Yang/PRISM.git}.
\endgroup
}

\section{Introduction}
Cities are increasingly instrumented with sensing and digital infrastructures that continuously generate heterogeneous data from mobility networks, land-use registries, transportation systems, and socio-economic activities~\cite{xu2025using}.
This data provides unprecedented opportunities for understanding urban dynamics. Yet, heterogeneity in terms of structure, granularity, noise ratio, and spatial coverage poses fundamental data integration and computational modeling challenges.

Urban region representation learning has emerged as a principled abstraction for addressing these challenges~\cite{DBLP:journals/pvldb/Duan0Z0WZLZR23,xu2025using,DBLP:journals/corr/abs-2609-15305,DBLP:conf/icde/CURE}.
Transforming diverse urban signals into compact embeddings enables computational models that capture latent spatial functionality and inter-region relationships in a unified vector form.
These embeddings serve as reusable, transferable representations that support a wide range of analytical tasks, including check-in prediction~\cite{DBLP:journals/tnn/LiangZ23,DBLP:journals/tkde/GongWGLLZWZL24,DBLP:journals/tkde/GongGLLZSLHW25}, crime forecasting~\cite{DBLP:journals/tcss/ButtLAS25,DBLP:conf/aaai/WangLYSYS22,DBLP:conf/icde/LiHXXP22,DBLP:conf/aaai/ZhaoFLT22,DBLP:journals/tkde/ZhaoLCZ23}, traffic modeling~\cite{DBLP:conf/cvpr/ZhangKID0I025,DBLP:conf/www/SunYLYS025,DBLP:journals/corr/abs-2511-20729,DBLP:journals/corr/abs-2106-09373,DBLP:conf/icde/YangGHYTJ22,DBLP:conf/kdd/YangHGYJ23,DBLP:conf/kdd/YangSHXTLYJ26,DBLP:journals/corr/abs-2609-15305,DBLP:conf/icde/CURE,DBLP:conf/ijcai/YangGHT021,DBLP:conf/CIKM/TFM,DBLP:journals/tkde/YangGY22,DBLP:conf/WWW/Path-LLM,DBLP:conf/KDD/MM-Path,DBLP:conf/icde/Yang020}, and service demand estimation~\cite{DBLP:conf/icde/Sun0CFKT24,DBLP:conf/kdd/SunCTK025}.

Given the heterogeneous nature of urban data, a central question arises:
\emph{how should multiple urban signals be integrated to construct reliable region representations?}

\noindent\textbf{Structured multi-view fusion.}
Existing approaches treat each urban modality—such as mobility flow graphs, POI distributions, or land-use maps—as an independent structured signal that is encoded using graph or attribute-based models.
Representative methods (e.g., \texttt{MGFN}~\cite{DBLP:conf/ijcai/WuYFPZZ0W22}, \texttt{HAFusion}~\cite{DBLP:conf/icde/Sun0CFKT24}, \texttt{CGAP}~\cite{DBLP:conf/ijcai/XuZ24}) jointly learn representations from multiple modalities and fuse them by means of concatenation, attention, or weighted aggregation.
More recent studies extend this paradigm to cross-modal settings by incorporating additional sources such as street-view images (e.g., \texttt{FlexiReg}~\cite{DBLP:conf/kdd/SunCTK025}).
While being architecturally diverse, these methods adopt a shared abstraction: heterogeneous views are integrated via globally shared fusion mechanisms under deterministic parameterization.

\noindent\textbf{Multi-objective optimization in multi-view learning.}
Multi-view urban representation learning typically introduces modality-specific objectives that guide the shared embedding towards distinct structural or semantic properties.
These objectives are then jointly optimized within a unified parameter space without explicit coordination.
Although general multi-task strategies (e.g., gradient normalization or projection) have been explored in other domains, they assume task symmetry and do not account for the spatially heterogeneous and modality-dependent dynamics intrinsic to urban data.

Despite their methodological variations, existing approaches rely on two implicit assumptions:
\textit{(i) modality importance is spatially consistent across regions}, and
\textit{(ii) joint optimization of heterogeneous objectives naturally produces coordinated and balanced representations}.

\begin{figure}[t]
	\centering
	\includegraphics[width=\linewidth]{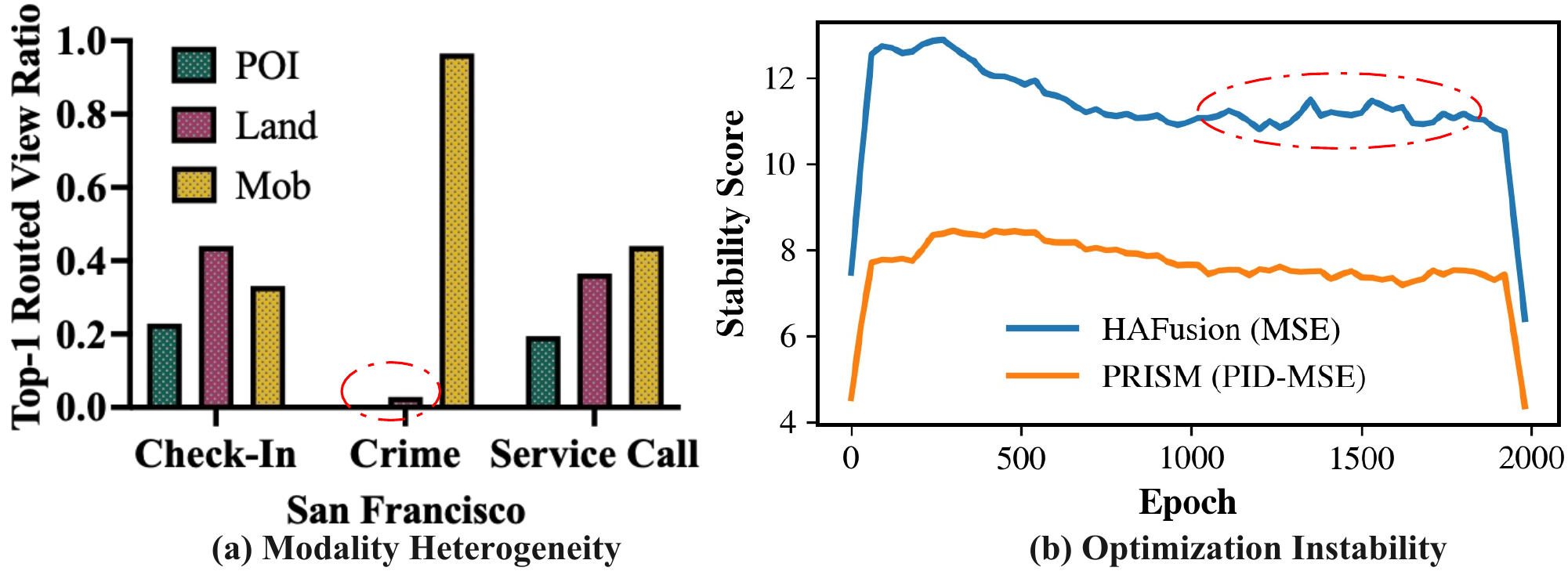}
	\caption{Motivation Example on San Francisco Dataset.}
	\label{fig:me}
	\vspace{-20pt}
\end{figure}

\noindent\textbf{Motivating example.}
To evaluate these assumptions, we analyze three city-scale datasets (see Figure~\ref{fig:me}).
Figure~\ref{fig:me}a highlights task-dependent differences in modality utility: mobility signals are most informative for crime prediction, whereas land-use and POI features are more useful for service call prediction, suggesting differing underlying spatial dynamics. Notably, as highlighted by the red dashed oval, POI and land-use features contribute negligibly to the SF Crime task. This pronounced heterogeneity challenges the assumption of globally uniform fusion across modalities.
Figure~\ref{fig:me}b further reveals a systematic optimization imbalance under conventional joint training (e.g., the red dashed oval). Gradients associated with different modality-specific objectives exhibit persistent magnitude disparities, allowing dominant modalities to steer the shared embedding space while suppressing weaker yet informative signals. Such imbalance induces representational bias and ultimately degrades cross-region generalization.
Following this analysis, we identify two coupled research challenges that remain unresolved in urban region representation learning:

\noindent
\textbf{Challenge~1 (C1): Spatially adaptive view integration.}
The predictive utility of each modality is inherently region-dependent, and yet existing fusion mechanisms assign contributions through globally shared parameters. This mismatch between local data characteristics and global fusion strategies incurs representational bias—commercial regions may be well-served while residential regions may be systematically under-represented. The challenge is to enable \emph{region-aware, adaptive modality selection} that aligns representation capacity with spatial heterogeneity, without requiring region-specific supervision.

\noindent	
\textbf{Challenge~2 (C2): Coordinated optimization under heterogeneous objectives.}
When multiple modality-specific objectives with different gradient scales, noise levels, and convergence rates are jointly optimized in a shared embedding space, objectives associated with certain modalities may contribute disproportionately large or more stable gradients, causing them to dominate the optimization process and suppress weaker modality signals, which in turn leads to imbalanced representations and unstable training. The challenge is to \emph{explicitly regulate the interplay among competing objectives} so that all informative modalities contribute proportionally to the final representation.

To address the two challenges, we propose \texttt{PRISM} (\textbf{P}er-\textbf{R}egion \textbf{I}ntegration with \textbf{S}elf-regulating \textbf{M}ulti-view coordination), a unified framework that redefines multi-view urban representation learning as an \emph{adaptive coordination problem} rather than as a static aggregation procedure.
In particular, \texttt{PRISM} builds on an existing multi-view encoding backbone and focuses on two components: region-level sparse view routing and PID-inspired discrepancy regulation.

\noindent
\textbf{Region-level adaptive routing via sparse MoE.} For modeling, \texttt{PRISM} instantiates region-level view allocation with a standard sparse MoE routing design. Our goal is not to introduce a new generic MoE routing operator, but to adapt sparse routing to heterogeneous urban view fusion.
Different from token-level MoE routing over homogeneous sequence elements, the routed unit in \texttt{PRISM} is a region-view representation derived from mobility, POI, or land-use signals.
The router dynamically selects and prioritizes informative views for each region, enabling localized view specialization while suppressing redundant or noisy signals.
This addresses \textbf{C1} by aligning view contributions with the spatial heterogeneity observed in Figure~\ref{fig:me}(a).

\noindent
\textbf{PID-inspired multi-objective discrepancy regulation.}
For optimization, \texttt{PRISM} introduces a PID-inspired discrepancy regularizer to coordinate heterogeneous view-induced objectives.
We borrow the proportional--integral--derivative decomposition to model complementary discrepancy signals during training.
Specifically, the proportional term captures the current discrepancy, the integral term captures accumulated bias, and the derivative term captures short-term variation.
These terms are used as loss-level regularization penalties rather than as a single control signal acting on the system.
This transforms multi-view learning from implicit joint optimization into an explicitly regularized coordination process, addressing the gradient imbalance identified in \textbf{C2}.

In summary, the main contributions are as follows:

\begin{itemize}[leftmargin=*]
	\item \textbf{Region-Level Adaptive View Allocation.}
	We formulate multi-view urban representation learning as a region-level adaptive view allocation problem, where different regions may rely on different subsets of mobility, POI, and land-use signals.
	We instantiate this formulation with sparse MoE routing to enable selective view integration at the region granularity.
	
	\item \textbf{PID-Inspired Optimization Coordination.}
	We formulate multi-view representation learning as an adaptive optimization coordination problem and introduce a PID-inspired discrepancy regularizer that captures current discrepancy, accumulated bias, and local variation.
	The final loss aggregation adopts uncertainty-based weighting to adaptively balance heterogeneous objectives, while the proposed contribution lies in PID-inspired discrepancy modeling rather than the weighting scheme.
	This design mitigates gradient conflicts and stabilizes training under heterogeneous view-induced objectives.
	
	\item \textbf{Comprehensive Empirical Validation.}
	Extensive experiments on three large-scale urban datasets across multiple downstream tasks evaluate the effectiveness, robustness, and efficiency of \texttt{PRISM}.
\end{itemize}

\section{Related Work}

\subsection{Urban Region Representation Learning}
Urban region representation learning aims to derive compact embeddings that encode functional, structural, and interaction patterns of spatial regions. From a data modeling perspective, existing work can be broadly categorized into \textit{single-source modeling} and \textit{heterogeneous multi-source modeling}, according to how urban signals are organized and integrated.

Single-source approaches learn region embeddings from a dominant urban signal~\cite{DBLP:conf/ijcai/YaoFLHX18,DBLP:conf/ijcai/WuYFPZZ0W22}, such as mobility flows, POI distributions, or land-use attributes. For example \texttt{ZE-Mob}~\cite{DBLP:conf/ijcai/YaoFLHX18} encodes a mobility transition graph using skip-grams. This type of method typically constructs region-level graphs or attribute matrices and apply graph representation learning, random-walk-based embedding, or matrix factorization techniques to obtain compact representations. Some studies extend this paradigm by extracting multiple complementary views from the same modality, e.g., using temporal slicing of mobility flows or structural and semantic similarity graphs, and then aggregate these views, the goal being to enhance robustness~\cite{DBLP:conf/aaai/ZhouHCS023,DBLP:conf/ijcai/0004LLH20}. While such designs enrich representations based on a single modality, the learned embeddings remain shaped by the inductive bias and information scope of the specific modality.

To increase representational completeness, more recent studies~\cite{DBLP:conf/icde/Sun0CFKT24,DBLP:conf/kdd/LiHCW023,DBLP:conf/kdd/SunCTK025} integrate heterogeneous urban signals, including mobility, POIs, land use, socio-economic attributes, and environmental features. These methods seek to capture complementary information from different modalities by constructing multi-graph architectures, cross-attention mechanisms, shared latent spaces, or fusion-based embedding frameworks. Employing joint optimization, they aim to produce unified region representations that reflect multiple aspects of urban dynamics. Modality interactions are typically modeled through feature concatenation or shared attention mechanisms, resulting in a single embedding embodying information from all available views.

\subsection{Mixture-of-Experts and Adaptive Routing}

Mixture-of-experts (MoE) models enable conditional computation by routing inputs to specialized experts through learned gating functions~\cite{DBLP:conf/emnlp/LiSYJWX23,DBLP:conf/nips/NguyenHR24}.
They have been widely adopted in large-scale neural models to increase model capacity while maintaining computational efficiency~\cite{DBLP:journals/pacmmod/NieMWYXMC023,DBLP:conf/nips/LuZ0H0X24}.

Existing MoE routing strategies mainly differ in sparsity and expert-allocation design.
\texttt{Switch Transformer}~\cite{DBLP:journals/jmlr/FedusZS22} performs sparse token-level routing by assigning each token to selected feed-forward experts, while \texttt{GShard}~\cite{DBLP:conf/iclr/LepikhinLXCFHKS21} adopts top-$k$ routing with auxiliary balancing losses.
\texttt{Expert Choice}~\cite{DBLP:conf/nips/ZhouLLDHZDCLL22} reverses the assignment direction to improve expert utilization, and \texttt{Soft MoE}~\cite{DBLP:conf/iclr/PuigcerverRMH24} replaces discrete routing with continuous mixture weights.
These methods are primarily developed for homogeneous token streams and emphasize scalability, load balancing, and efficient conditional computation.

Recent studies extend MoE to multi-modal and spatio-temporal settings.
In multi-modal learning, methods such as \texttt{MoME}~\cite{DBLP:conf/nips/ShenCSGN24} and \texttt{VLMo}~\cite{DBLP:conf/nips/BaoW0LMASPW22} activate experts according to modality or layer-level signals.
In spatio-temporal forecasting, \texttt{ST-MoE}~\cite{DBLP:conf/cikm/LiCZYZ023} routes temporal inputs to specialized experts to capture temporal heterogeneity.
MoE has also been explored in urban computing. For example, \texttt{UrbanMoE}~\cite{DBLP:conf/www/LiuLZMJLZK26} applies sparse multi-modal expert routing to multi-task urban region profiling, using satellite imagery, textual descriptions, and POI features to predict multiple urban indicators.

Different from these studies, \texttt{PRISM} focuses on reusable urban region representation learning from mobility, POI, and land-use views. Its goal is not to introduce a new generic MoE routing operator, but to adapt standard sparse MoE routing to region-level heterogeneous view allocation. In this setting, the routed unit is a region-view representation, and the routing decision determines which urban views should contribute to each region. This differs from token-level routing in large language models, modality-aware routing in multi-modal learning, temporal routing in forecasting, and task-aware expert routing in urban profiling.

Recent meta-learning studies further explore spatial adaptation in urban prediction, such as region-adaptive prediction with region profiles and few-shot transfer from data-sufficient cities to data-scarce cities~\cite{DBLP:conf/cikm/ChenLL23,DBLP:conf/icpads/SunDZLXH23}. These methods mainly adapt prediction models under limited observations, whereas \texttt{PRISM} studies region-level coordination among structurally distinct urban views within a unified representation learning framework.

\subsection{Adaptive Multi-Task Learning and PID-Inspired Regulation}

Adaptive multi-task learning (MTL) jointly optimizes multiple objectives while mitigating task imbalance and gradient interference. Representative methods include \texttt{GradNorm}~\cite{DBLP:conf/icml/ChenBLR18}, which balances task training rates through gradient magnitude normalization; \texttt{PCGrad}~\cite{DBLP:conf/nips/YuK0LHF20}, which projects conflicting gradients; and \texttt{MGDA}~\cite{MGDA,DBLP:conf/nips/SenerK18}, which seeks a common descent direction toward Pareto-stationary solutions. These methods mainly operate on instantaneous task gradients and are typically designed for conventional MTL settings with comparable supervised tasks.

Several recent studies incorporate PID principles into neural learning systems, but at different levels. \texttt{PIDformer}~\cite{DBLP:conf/icml/NguyenUNB24} introduces a PID closed-loop feedback signal into transformer state-space dynamics to improve robustness and mitigate rank collapse. PID-based stochastic optimization~\cite{DBLP:conf/cvpr/AnWSXDZ18} treats gradients as control errors and updates network parameters using current, accumulated, and changing gradient information.

Different from these methods, \texttt{PRISM} addresses heterogeneous view-induced objectives in urban representation learning, where mobility models origin--destination interactions, while POI and land-use preserve view-specific similarity structures. \texttt{PRISM} does not apply PID to attention dynamics or optimizer states. Instead, it introduces a PID-inspired discrepancy regularizer over view-induced objectives, where the proportional, integral, and derivative terms model current discrepancy, accumulated bias, and short-term variation during training. We further include \texttt{GradNorm}, \texttt{PCGrad}, and \texttt{MGDA} as controlled baselines under the same backbone to empirically examine this distinction.

\begin{figure*}[t]
	\centering
	\includegraphics[scale=0.58]{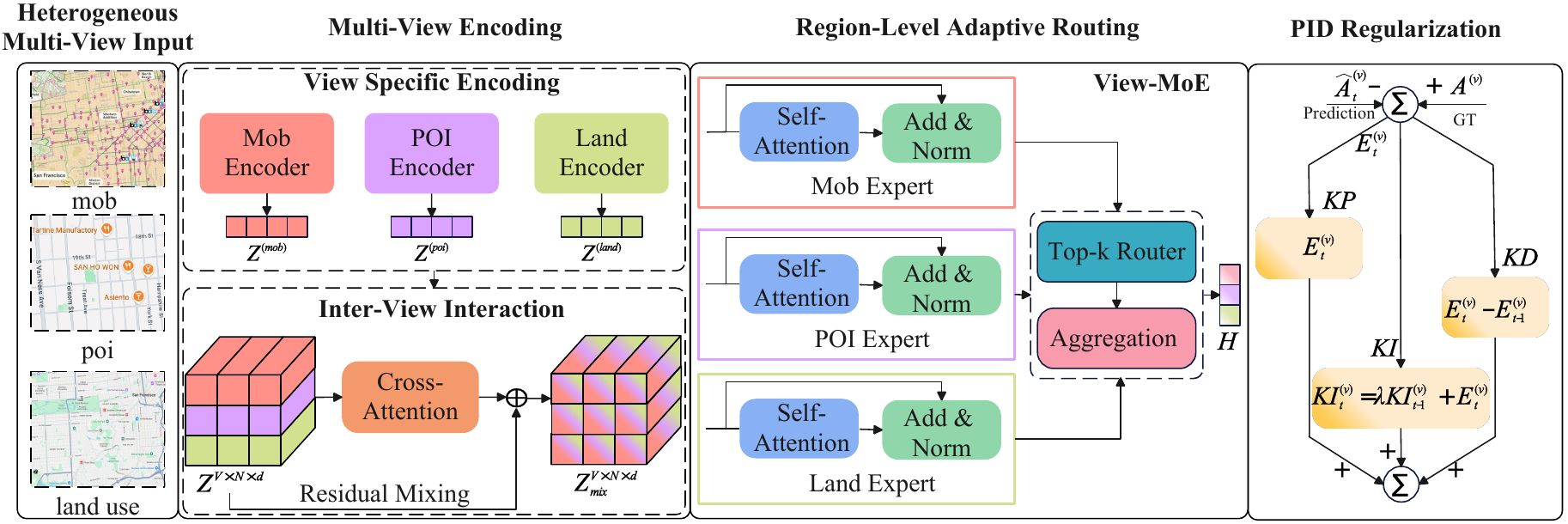}
\caption{The \texttt{PRISM} framework. The multi-view encoding block follows the \texttt{HAFusion}-style backbone; \texttt{PRISM} introduces region-level adaptive routing and PID-inspired discrepancy regularization. GT denotes ground truth.}
	\label{fig:framework}
	\vspace{-10pt}
\end{figure*}	
\section{Preliminaries}
We first cover important concepts that underlie the paper's proposal and then state the problem addressed.

\subsection{Definitions}
\begin{myDef}[Region]
	Given a geographical area $\mathcal{A}$, we define a region set 
	$\mathcal{R} = \{\mathcal{R}_1, \dots, \mathcal{R}_N\}$ 
	as a spatial partition of $\mathcal{A}$ such that 
	$\mathcal{R}_i \cap \mathcal{R}_j = \emptyset$ for $i \neq j$ 
	and $\bigcup_{i=1}^{N} \mathcal{R}_i = \mathcal{A}$. For each region, we consider three types of heterogeneous features to construct its representation.
\end{myDef}

\begin{myDef}[Human Mobility Feature]
	
	Given a region set $\mathcal{R}=\{\mathcal{R}_1,\dots,\mathcal{R}_n\}$,
	a human mobility feature is a weighted directed graph 
	$\mathcal{G}_M = (\mathcal{R}, \mathbf{M})$, 
	where $\mathbf{M} \in \mathbb{R}^{n \times n}$ is a mobility matrix and 
	$m_{i,j}$ denotes the flow volume from $\mathcal{R}_i$ to $\mathcal{R}_j$
	within an observation period.
	Given a region $\mathcal{R}_i$, its mobility feature vector $m_i$ is defined as
	its outgoing flow vector $ \mathbf{M}_{i,:}$.
\end{myDef}

\begin{myDef}[POI Feature]
	
	Given a region set $\mathcal{R}=\{\mathcal{R}_1,\dots,\mathcal{R}_n\}$,
	a point-of-interest (POI) feature is a matrix
	$\mathbf{P} \in \mathbb{R}^{n \times d_P}$,
	where $d_P$ denotes the number of POI categories.
	Each entry $p_{i,k}$ records the frequency (or normalized count) 
	of POIs in $\mathcal{R}_i$ belonging to category $k$.
	The POI feature vector $\mathbf{p}_i$ of region $\mathcal{R}_i$ is defined as
	$\mathbf{P}_{i,:} \in \mathbb{R}^{d_P}$.
\end{myDef}

\begin{myDef}[Land-Use Feature]
	
	Given a region set $\mathcal{R}=\{\mathcal{R}_1,\dots, \\ \mathcal{R}_n\}$,
	a land-use feature is a matrix
	$\mathbf{L} \in \mathbb{R}^{n \times d_L}$,
	where $d_L$ denotes the number of land-use categories.
	Each entry $l_{i,k}$ records the frequency (or proportion) 
	of land parcels in $\mathcal{R}_i$ that belong to category $k$.
	The land-use feature vector $\mathbf{l}_i$ of region $\mathcal{R}_i$ is defined as
	$ \mathbf{L}_{i,:} \in \mathbb{R}^{d_L}$.
\end{myDef}

\begin{myDef}[Heterogeneous Multi-View Feature]
	
	Given a region set $\mathcal{R}=\{\mathcal{R}_1,\dots,\mathcal{R}_n\}$,
	we associate each region with $V$ heterogeneous feature views:
	\begin{equation}
	\mathcal{X} = \{\mathbf{X}^{(1)}, \mathbf{X}^{(2)}, \dots, \mathbf{X}^{(V)}\},
	\end{equation}
	where each view-specific feature matrix 
	$\mathbf{X}^{(v)} \in \mathbb{R}^{n \times d_v}$
	represents region-level attributes under view $v$.
	For a region $\mathcal{R}_i$, its feature vector under view $v$ is
	$\mathbf{x}_i^{(v)} = \mathbf{X}^{(v)}_{i,:} \in \mathbb{R}^{d_v}$. In our setting, $V=3$, corresponding to mobility, POI, and land-use views.
\end{myDef}

\begin{myDef}[Gradient Stability Score]
	\label{def:stability_energy}
	Let $\{\mathcal{L}_v(\theta)\}_{v=1}^{V}$ denote modality-specific objectives
	sharing parameters $\theta$.
	At the $t$-th optimization step, let $g_t^{(v)}=\nabla_\theta \mathcal{L}_v(\theta_t)$.
	We define the gradient stability score as follows:
	\begin{equation}
		\label{eq:ss}
		S_t
		=
\log
\frac{\max_{v} \|g_t^{(v)}\|+\epsilon_g}
{\min_{v} \|g_t^{(v)}\|+\epsilon_g}
		+
		\lambda_c
		\frac{1}{|\mathcal{P}|}
		\sum_{(i,j)\in\mathcal{P}}
		\max\!\left(0,-\cos\!\left(
		g_t^{(i)}, g_t^{(j)}
		\right)\right),
	\end{equation}
	where $\epsilon_g>0$ is a small numerical floor that prevents the ratio from becoming unbounded when a modality gradient vanishes. $\mathcal{P}$ denotes all unordered modality pairs
	and $\lambda_c>0$ balances gradient magnitude imbalance
	and gradient conflict. Intuitively, the first term measures gradient magnitude imbalance,
	while the second term captures directional conflicts across modalities.
	A smaller $S_t$ indicates more balanced and aligned multi-objective optimization. When all modality gradients have identical norms and
	non-negative pairwise cosine similarity,
	both imbalance and conflict terms vanish,
	resulting in $S_t=0$. Moreover, any update rule that reduces gradient norm disparity
	or alleviates pairwise conflicts decreases the stability score monotonically.
	Therefore, $S_t$ serves as a principled metric
	for quantifying improvements in gradient coordination.
\end{myDef}

\begin{myDef}[Self-Attention]
	Given an input matrix $\mathbf{X} \in \mathbb{R}^{n \times d}$,
	self-attention computes contextualized representations by modeling pairwise interactions among its elements.
	Let
	\begin{equation}
	\mathbf{Q} = \mathbf{X}\mathbf{W}_Q,\quad
	\mathbf{K} = \mathbf{X}\mathbf{W}_K,\quad
	\mathbf{V} = \mathbf{X}\mathbf{W}_V,
	\end{equation}
	where $\mathbf{W}_Q,\mathbf{W}_K,\mathbf{W}_V \in \mathbb{R}^{d \times d}$ are learnable parameters.
	The self-attention operation is defined as follows:
	\begin{equation}
		\mathrm{Attn}(\mathbf{X})
		=
		\mathrm{Softmax}\!\left(
		\frac{\mathbf{Q}\mathbf{K}^\top}{\sqrt{d}}
		\right)\mathbf{V}.
	\end{equation}
\end{myDef}

\subsection{Problem Statement}

Given a set of urban regions $\mathcal{R}=\{R_1,\ldots,R_N\}$ and $V$ heterogeneous urban views 
	$\mathcal{X}=\{X^{(1)},\ldots,X^{(V)}\}$, where 
	$X^{(v)}\in\mathbb{R}^{N\times d_v}$ denotes the feature matrix of view $v$, our goal is to learn a unified region representation matrix 
	$H\in\mathbb{R}^{N\times d}$ for downstream urban prediction tasks.

Unlike conventional multi-view fusion that applies a fixed fusion rule to all regions, we aim to learn region-adaptive representations. Specifically, each region $R_i$ should be allowed to assign different importance to different views:
	\begin{equation}
		h_i = \Phi\left(\sum_{v=1}^{V}\alpha_{i,v} z_i^{(v)}\right),
		\quad
		\sum_{v=1}^{V}\alpha_{i,v}=1,\quad
		\alpha_{i,v}\geq 0,
	\end{equation}
	where $z_i^{(v)}$ is the view-specific embedding of region $R_i$, and $\alpha_{i,v}$ denotes the region-specific contribution of view $v$. This formulation captures the spatial heterogeneity of urban regions: different regions may depend on different views due to their distinct functional, mobility, and land-use characteristics.

\section{Methodology}
Figure~\ref{fig:framework} presents the overall architecture of \texttt{PRISM}.
The multi-view encoding component follows the \texttt{HAFusion}-style backbone and is used to encode mobility, POI, and land-use data defined over shared spatial regions.
On top of this backbone, \texttt{PRISM} introduces region-level adaptive routing to allocate view contributions for each region.
It further incorporates PID-inspired discrepancy regularization to mitigate gradient imbalance and cross-objective conflicts among heterogeneous view-induced objectives.
\subsection{Multi-View Encoding}

\noindent
\textbf{View-Specific Encoding.}
Given heterogeneous urban features, we first map each modality into a shared latent space of dimensionality $d$.
We employ an MLP (multilayer perceptron) layer to project POI and land-use features into $d$-dimensional embeddings,
whereas mobility features are inherently represented in the same embedding space.
Let $\mathbf{X}^{(v)} \in \mathbb{R}^{N \times d}$ denote the initial embedding of modality 
$v \in \{\text{mob}, \text{poi}, \text{land}\}$ over $N$ regions.

To capture modality-specific structural dependencies,
we use an intra-view encoder based on stacked self-attention layers:
\begin{equation}
	\mathbf{Z}^{(v)} = \mathrm{IntraEncoder}(\mathbf{X}^{(v)}),
\end{equation}

Within each modality, multi-head self-attention is applied to capture long-range regional dependencies.
This step ensures that the intrinsic topological and semantic properties of each view are retained prior to cross-modal fusion.

\medskip
\noindent
\textbf{Inter-View Interaction.}
While intra-view encoding captures moda\\lity-specific structural characteristics, urban functionality typically arises from the interaction among heterogeneous signals.
To facilitate cross-modal interaction, we organize modality embeddings into a unified representation framework:
\begin{equation}
\mathbf{Z} \in \mathbb{R}^{V \times N \times d},
\end{equation}
where $V$ denotes the number of modalities.
For each region, we apply an inter-view attention mechanism to learn cross-modal correlations among modality-specific embeddings:
\begin{equation}
	\tilde{\mathbf{Z}} = \mathrm{InterEncoder}(\mathbf{Z}),
\end{equation}

\medskip
\noindent
\textbf{Residual Mixing.}
To balance modality-specific structural signals and cross-view interactions, we introduce learnable mixing coefficients $\alpha$ and $\beta$:
\begin{equation}
	\mathbf{Z}^{(v)}_{\text{mix}}
	=
	\beta \tilde{\mathbf{Z}}^{(v)}
	+
	\alpha \mathbf{Z}^{(v)}.
\end{equation}
This residual formulation preserves stable intra-view representations while adaptively integrating cross-view information, thereby providing a flexible basis for subsequent adaptive fusion.

\subsection{Region-Level Adaptive Routing}
\label{sec:rar}
This module instantiates the standard sparse MoE routing paradigm for region-level heterogeneous view allocation. We do not aim to introduce a new generic routing operator; rather, we adapt learned gating and top-$k$ sparse selection to the urban multi-view setting, where each routing decision determines which view should contribute to a specific region.

Given the mixed multi-view representations
$\{\mathbf{Z}^{(v)}_{\mathrm{mix}}\}_{v=1}^{V}$, where
$\mathbf{Z}^{(v)}_{\mathrm{mix}} \in \mathbb{R}^{N \times d}$,
we perform region-level adaptive fusion to integrate heterogeneous views.

\medskip
\noindent
\textbf{Routing weight computation.}
For region $i$ and view $v$, let
$\mathbf{Z}^{(v)}_{\mathrm{mix},i} \in \mathbb{R}^{d}$
denote the corresponding mixed representation.
A lightweight gating network computes a routing logit as follows:
\begin{equation}
	s_{v,i}
	=
	\mathbf{w}_2^\top \sigma
	\!\left(
	\mathbf{W}_1 \mathbf{Z}^{(v)}_{\mathrm{mix},i} + \mathbf{b}_1
	\right)
	+ b_2,
\end{equation}
where $\mathbf{W}_1$, $\mathbf{w}_2$, $\mathbf{b}_1$, and $b_2$ are learnable parameters, and $\sigma(\cdot)$ is the ReLU activation function.
Let $\mathcal{T}_i$ denote the indices of the top-$k$ views for region $i$ according to the logits $\{s_{v,i}\}_{v=1}^{V}$. The routing weight is computed by a masked softmax:
\begin{equation}
	\alpha_{v,i}
	=
	\frac{\mathbb{I}(v\in\mathcal{T}_i)\exp(s_{v,i}/\tau)}
	{\sum_{v'\in\mathcal{T}_i}\exp(s_{v',i}/\tau)} ,
\end{equation}
where $\tau>0$ controls the sharpness of the routing distribution.
When collecting all routing weights, we obtain a routing matrix
$\boldsymbol{\alpha} \in \mathbb{R}^{V \times N}$.

\medskip
\noindent
\textbf{Adaptive fusion.}
The fused representation of region $i$ is given by:
\begin{equation}
	\mathbf{H}_i
	=
	\sum_{v=1}^{V}
	\alpha_{v,i}\mathbf{Z}^{(v)}_{\mathrm{mix},i}.
\end{equation}

\medskip
\noindent
\textbf{Routing sparsity regularization.} To encourage region-level selective view allocation, we introduce an optional entropy-based sparsity regularizer:
	\begin{equation}
		\mathcal{L}_{\mathrm{sparse}}
		=
		\frac{1}{N}
		\sum_{i=1}^{N}
		\mathcal{H}(\boldsymbol{\alpha}_i),
	\end{equation}
	where
	$\boldsymbol{\alpha}_i = [\alpha_{1,i}, \dots, \alpha_{V,i}]^\top$
	is the routing distribution of region $i$, and $\mathcal{H}(\cdot)$ denotes Shannon entropy.
	Minimizing $\mathcal{L}_{\mathrm{sparse}}$ encourages sharper region-specific view selection and reduces overly uniform cross-view mixing.

\subsection{PID-Inspired Discrepancy Regularization}
\label{sec:pid_theory}

While the proposed routing mechanism enables flexible multi-view integration,
the resulting optimization process may still suffer from gradient imbalance and oscillatory discrepancy trajectories.
Inspired by the proportional--integral--derivative decomposition in control theory, we introduce a PID-inspired discrepancy regularizer.
We emphasize that this formulation is not a classical PID controller: instead of combining P/I/D terms into a single control signal, it penalizes the squared proportional, accumulated, and differential discrepancies as separate loss components.
Thus, PID-MSE should be understood as a loss-level regularization objective rather than a structurally equivalent PID control law.

\noindent\textbf{PID-MSE Formulation. }
Let $\theta$ denote the shared backbone parameters.
	PID-MSE is applied only to similarity-based views, i.e., $v\in\{\mathrm{poi},\mathrm{land}\}$.
	For these views, we define the discrepancy as:
\begin{equation}
E_t^{(v)} = \hat{A}_t^{(v)} - A^{(v)},
\end{equation}
where $A^{(v)}$ denotes the target region--region similarity matrix constructed from view-specific attributes, and $\hat{A}_t^{(v)}$ denotes the predicted similarity matrix at training step $t$.
Instead of minimizing only the instantaneous discrepancy $\|E_t^{(v)}\|_2^2$, as in standard mean squared error (MSE) objectives, we introduce a PID-inspired regularization term that incorporates proportional, integral, and derivative components: 
\begin{align}
	P_t^{(v)} &= E_t^{(v)}, 
	I_t^{(v)} = \lambda I_{t-1}^{(v)} + E_t^{(v)}, \\
	D_t^{(v)} &= E_t^{(v)} - E_{t-1}^{(v)},
\end{align}
We then define the PID-regularized objective as follows:
\begin{equation}
	\mathcal{L}_{\mathrm{PID}}^{(v)}
	=
	\alpha\|P_t^{(v)}\|_2^2
	+
	\beta \|I_t^{(v)}\|_2^2
	+
	\gamma\|D_t^{(v)}\|_2^2,
\end{equation}
where $\alpha>0$, $\beta,\gamma\geq 0$, and $\lambda\in(0,1]$.
When $\beta=\gamma=0$, PID-MSE reduces to the standard MSE objective.

The proportional term enforces instantaneous alignment,
the integral term captures accumulated discrepancies over time,
and the derivative term penalizes rapid discrepancy variations.
Together, these terms reshape the optimization dynamics by
stabilizing gradient trajectories and mitigating oscillatory updates.
From a multi-objective perspective, PID-MSE exposes the optimization process to instantaneous, accumulated, and differential discrepancy information, which can help reduce gradient imbalance and oscillatory updates.
This intended regularization effect is empirically examined by the stability analysis in Fig.~\ref{fig:sub2} and Section~\ref{sec:mba}.

\noindent\textbf{Stability Analysis}
We next analyze how each term of the PID objective influences the optimization dynamics.
Specifically, we provide a local optimization interpretation showing how the proportional term encourages discrepancy reduction, the derivative term penalizes rapid discrepancy changes, and the integral term captures persistent bias.
Together, these effects explain the empirical stability improvements observed in Fig.~1(b).

\begin{proposition}[Proportional Term $P$: Local Discrepancy Reduction]
	\label{prop:p_term}
	Assume that $E(\theta)$ is continuously differentiable and locally $L$-smooth.
	For a sufficiently small learning rate $\eta$,
	minimizing $\alpha\|E_t\|_2^2$ ensures
	$
	\|E_{t+1}\|_2 \le (1 - c\eta)\|E_t\|_2
	$
	for some $c>0$.
\end{proposition}

\paragraph{Proof sketch.}
Since $\nabla_\theta \|E_t\|_2^2 = 2J_t^\top E_t$,
the update direction aligns with the instantaneous discrepancy.
Under local smoothness and small $\eta$,
this induces a contraction mapping on $E_t$.
\qed
\begin{proposition}[Derivative Term $D$: Oscillation Suppression]
	\label{prop:d_term}
	Under the same assumptions,
	augmenting the objective with $\gamma\|E_t - E_{t-1}\|_2^2$ ($\gamma>0$)
	reduces the upper bound of discrepancy variation
	$\|E_{t+1} - E_t\|_2$.
\end{proposition}

\paragraph{Proof sketch.}
The derivative regularizer introduces a gradient component
proportional to $(E_t - E_{t-1})$,
penalizing rapid discrepancy changes.
This produces a damping effect in the discrepancy trajectory,
thereby suppressing oscillatory updates.
\qed
\begin{proposition}[Integral Term $I$: Bias Correction]
	\label{prop:i_term}
	If the discrepancy exhibits a persistent non-zero average,
	then for $\beta>0$ and $\lambda\in(0,1]$,
	the integral state accumulates this bias
	and induces a sustained corrective gradient component.
\end{proposition}

\paragraph{Proof sketch.}
$I_t$ is an exponentially weighted accumulation of past discrepancies.
Persistent bias prevents cancellation,
so $\|I_t\|$ remains non-zero,
producing a non-vanishing corrective signal.
\qed

Notably, the proportional term ensures instantaneous error reduction,
the derivative term attenuates oscillatory updates,
and the integral term compensates for persistent bias~\cite{DBLP:conf/icml/NguyenUNB24}.
Collectively, these terms enhance the stability of multi-objective optimization,
which is particularly important in heterogeneous multi-view settings where competing modality gradients often occur.

\subsection{Training Objective}
\label{sec:train_obj}
Given the region embeddings $\mathbf{H}\in\mathbb{R}^{N\times d}$ produced by the preceding modules,
we define the overall optimization objective of \texttt{PRISM}.
The model is trained with three modality-specific losses,
together with an optional routing sparsity regularizer.
Let $\theta$ denote the collection of trainable parameters.

\noindent\textbf{Mobility objective.}
Following prior work on urban flow modeling~\cite{DBLP:journals/pvldb/FangPCDG21,DBLP:journals/pvldb/MaWWZWW25,DBLP:journals/pvldb/HanWXL25},
we reconstruct the observed mobility matrix $\mathbf{M}\in\mathbb{R}^{N\times N}$
via a bidirectional softmax likelihood.
Let $\mathbf{H}^s$ and $\mathbf{H}^t$ denote projected embeddings of $\mathbf{H}$ for source and target, respectively.
The mobility loss is defined as follows:
\begin{equation}
	\mathcal{L}_{\text{mob}}
	=
	-\sum_{i,j} M_{i,j}\log p_{i\rightarrow j}
	-\sum_{i,j} M_{j,i}\log p_{j\rightarrow i},
\end{equation}
where
$
p_{i\rightarrow j}
=
\mathrm{Softmax}_j\!\left(\mathbf{H}^s_i(\mathbf{H}^t_j)^\top\right),
$
and $\mathrm{Softmax}_j(\cdot)$ normalizes over all destination regions $j$.

\noindent\textbf{POI and land-use objectives.}
For attribute-based modalities, we aim to preserve region-level similarity structures.
For each modality $v\in\{\text{poi},\text{land}\}$,
we align the predicted similarity matrix $\hat{\mathbf{A}}^{(v)}$
with the corresponding target similarity matrix $\mathbf{A}^{(v)}$
by minimizing
\begin{equation}
	\mathcal{L}_{v}
	=
	\ell\!\left(\hat{\mathbf{A}}^{(v)}, \mathbf{A}^{(v)}\right),
	\quad
	v\in\{\text{poi},\text{land}\},
\end{equation}
where $\ell(\cdot)$ denotes the PID-MSE loss defined in Section~\ref{sec:pid_theory}.

\noindent
\textbf{Routing sparsity regularization.}
When region-level adaptive routing is enabled,
we encourage selective view allocation
by minimizing the entropy of routing distributions:
\begin{equation}
	\mathcal{L}_{\text{sparse}}
	=
	\frac{1}{N}\sum_{i=1}^{N}\mathcal{H}(\boldsymbol{\alpha}_i),
\end{equation}
where $\boldsymbol{\alpha}_i\in\mathbb{R}^{V}$ denotes the routing distribution of region $i$.
Minimizing entropy promotes sharper, more confident modality selection,
preventing overly uniform fusion and encouraging regions to give preference to the most informative views.

\noindent\textbf{Overall objective.}
The final training objective combines all modality-specific losses
together with routing regularization:
\begin{equation}
		\mathcal{L}
		=
		\mathcal{L}_{\text{mob}}
		+
		\mathcal{L}_{\text{poi}}
		+
		\mathcal{L}_{\text{land}}
		+
		\lambda_{\text{sparse}}\mathcal{L}_{\text{sparse}},
			\label{eq:full_obj}
\end{equation}
where $\lambda_{\text{sparse}}\geq 0$ controls the strength of region-level routing sparsity regularization.
To balance heterogeneous objectives, we adopt uncertainty-based weighting~\cite{DBLP:conf/cvpr/KendallGC18,DBLP:conf/cvpr/MaNLN025},
yielding the final optimized objective in Eq.~\ref{eq:full_obj}.

\subsection{Model Complexity Analysis}

Let $N$ denote the number of regions, $V$ the number of views, and $d$ the embedding dimensionality.
\texttt{HAFusion} and \texttt{PRISM} share identical intra-view and inter-view encoders,
whose dominant computational cost arises from self-attention operations,
resulting in a complexity of $O(VN^2d)$.
For region-level fusion, \texttt{HAFusion} adopts static weighted aggregation,
incurring $O(VNd)$ computation.
In contrast, \texttt{PRISM} introduces an additional routing module
with complexity $O(VNd h)$,
where $h \ll N$ denotes the router hidden dimensionality.
Since $h$ is typically much smaller than $N$,
the routing overhead remains negligible relative to the backbone attention cost.
Next, considering parameters,
\texttt{HAFusion} requires $O(Vd)$ fusion weights,
whereas \texttt{PRISM} adds $O(dh)$ router parameters.
Both are minor compared to the $O(d^2)$ parameters of Transformer blocks.
Therefore, although \texttt{PRISM} introduces additional routing parameters, they are lightweight and serve region-adaptive view selection rather than model scaling. By learning region-specific modality allocations, \texttt{PRISM} better captures spatial heterogeneity while maintaining the same asymptotic time and space complexity as \texttt{HAFusion}.

\begin{table*}[t]
	\centering
	\small
	\caption{Overall Performance on three datasets. Symbols '$\uparrow$' and '$\downarrow$' indicates that high values and small values are preferred, respectively. For each dataset, the best results are in bold/red, and the second-best results are underlined/in blue.}
	\begin{tabular}{l|l|lll|lll|lll}
		\toprule[1.5pt]
		\multirow{2}{*}{\textbf{}}             & \multicolumn{1}{c|}{\multirow{2}{*}{\textbf{Method}}} & \multicolumn{3}{c|}{\textbf{NYC}}                                            & \multicolumn{3}{c|}{\textbf{Chi}}                                                  & \multicolumn{3}{c}{\textbf{SF}}                                            \\ \cline{3-11} 
		& \multicolumn{1}{c|}{}                                  & \multicolumn{1}{c|}{\textbf{MAE} ($\downarrow$)} & \multicolumn{1}{c|}{\textbf{RMSE} ($\downarrow$)} & \multicolumn{1}{c|}{\textbf{$R^2$} ($\uparrow$)}  & \multicolumn{1}{c|}{\textbf{MAE} ($\downarrow$)} & \multicolumn{1}{c|}{\textbf{RMSE}$\downarrow$} &\multicolumn{1}{c|}{ \textbf{$R^2$} ($\uparrow$)}   & \multicolumn{1}{c|}{\textbf{MAE} ($\downarrow$)} & \multicolumn{1}{c|}{\textbf{RMSE} ($\downarrow$)} & \multicolumn{1}{c}{\textbf{$R^2$} ($\uparrow$)}   \\ \hline \hline
		\multirow{11}{*}{\textbf{\rotatebox{90}{Check-in}}}     & \texttt{MVURE}                                         & \multicolumn{1}{l|}{306.7±8.2}   & \multicolumn{1}{l|}{499.6±12.9}    & 0.627 ± 0.019 & \multicolumn{1}{l|}{1693 ± 74}    & \multicolumn{1}{l|}{3171+128}      & 0.656 ± 0.029 & \multicolumn{1}{l|}{346.8+8.7}    & \multicolumn{1}{l|}{659.3 ± 15.7}  & 0.562 ± 0.021 \\ \cline{2-11} 
		& \texttt{MGFN}                                          & \multicolumn{1}{l|}{292.6±17.1}   & \multicolumn{1}{l|}{451.8 ± 28.1}  & 0.690 ± 0.040 & \multicolumn{1}{l|}{1281 ± 41}    & \multicolumn{1}{l|}{2276 ± 86}    & 0.817 ± 0.011 & \multicolumn{1}{l|}{310.8 ± 9.1}  & \multicolumn{1}{l|}{542.1 ± 17.6}  & 0.708 ± 0.010 \\ \cline{2-11} 
		& \texttt{RDCL}                                     & \multicolumn{1}{l|}{371.2±10.3}   & \multicolumn{1}{l|}{495.5 ± 15.9}  & 0.471 ± 0.023 & \multicolumn{1}{l|}{2427 ± 123}   & \multicolumn{1}{l|}{4184 ± 136}     & 0.402 ± 0.042 & \multicolumn{1}{l|}{398.8  ± 9.9  }  & \multicolumn{1}{l|}{748.1 ± 17.8}  & 0.437 ± 0.024 \\ \cline{2-11} 
		& \texttt{HREP}                                          & \multicolumn{1}{l|}{276.3±11.7}   & \multicolumn{1}{l|}{448.2 ± 17.1}  & 0.703 ± 0.021 & \multicolumn{1}{l|}{1679 ± 71}    & \multicolumn{1}{l|}{3135 ± 79}    & 0.664 ± 0.017 & \multicolumn{1}{l|}{330.9 ± 9.3}  & \multicolumn{1}{l|}{606.7 ± 25.8}  & 0.629 ± 0.032 \\ \cline{2-11} 
		
		& \texttt{BiST}                                      & \multicolumn{1}{l|}{278.9 ± 4.7}   & \multicolumn{1}{l|}{442.6 ± 23.5}    & 0.707 ± 0.001 & \multicolumn{1}{l|}{1627.6 ± 32}   & \multicolumn{1}{l|}{2832 ± 49}   & 0.725 ± 0.002 & \multicolumn{1}{l|}{439.8 ± 8.3}  & \multicolumn{1}{l|}{704.9 ± 14.0}   & 0.497 ± 0.008 \\ \cline{2-11} 
		& \texttt{START}                                      & \multicolumn{1}{l|}{280.5 ± 8.9}   & \multicolumn{1}{l|}{458.9 ± 19.0}    & 0.685 ± 0.025 & \multicolumn{1}{l|}{1466 ± 45}   & \multicolumn{1}{l|}{3325 ± 136}   & 0.622 ± 0.031 & \multicolumn{1}{l|}{305.8 ± 7.2}  & \multicolumn{1}{l|}{645.9 ± 15.6}   & 0.580 ± 0.020 \\ \cline{2-11} 
		& \texttt{UrbanMoE}                                      & \multicolumn{1}{l|}{376.2 ± 8.2}   & \multicolumn{1}{l|}{639.8 ± 8.3}    & 0.388 ± 0.016 & \multicolumn{1}{l|}{2076 ± 78}   & \multicolumn{1}{l|}{4426 ± 120}   & 0.331 ± 0.036 & \multicolumn{1}{l|}{401.3 ± 7.8}  & \multicolumn{1}{l|}{820.1 ± 14.5}   & 0.323 ± 0.024 \\ \cline{2-11} 
		& \texttt{HGN}                                      & \multicolumn{1}{l|}{213.4 ± 9.9}    & \multicolumn{1}{l|}{337.6 ± 1.3}     & 0.829 ± 0.011 & \multicolumn{1}{l|}{1037 ± 25}  & \multicolumn{1}{l|}{2328 ± 29}  & 0.814 ± 0.042 & \multicolumn{1}{l|}{248.2 ± 3.6}  & \multicolumn{1}{l|}{442.5 ± 1.6}   & 0.802 ± 0.012 \\ \cline{2-11} 
		& \texttt{PGN}                                    & \multicolumn{1}{l|}{219.9 ± 3.0}    & \multicolumn{1}{l|}{363.3 ± 5.6}     & 0.803 ± 0.006 & \multicolumn{1}{l|}{964 ± 27}  & \multicolumn{1}{l|}{2160 ± 120}  & 0.841 ± 0.018 & \multicolumn{1}{l|}{243.1 ± 2.8}  & \multicolumn{1}{l|}{456.2 ± 8.0}   & 0.791 ± 0.007 \\ \cline{2-11}                                       
		& \texttt{HAF}                                      & \multicolumn{1}{l|}{\textcolor{blue}{\underline{202.8±7.2}}}    & \multicolumn{1}{l|}{\textcolor{blue}{\underline{322.8 ± 12.6}}}  & \textcolor{blue}{\underline{0.844 ± 0.012}} & \multicolumn{1}{l|}{\textcolor{blue}{\underline{929 ± 62}}}     & \multicolumn{1}{l|}{\textcolor{blue}{\underline{1947 ± 75}}}     &\textcolor{blue}{\underline{ 0.870 ± 0.010}} & \multicolumn{1}{l|}{\textcolor{blue}{\underline{233.1 ± 9.5 }}}  & \multicolumn{1}{l|}{\textcolor{blue}{\underline{429.6 ± 28.1}}}  & \textcolor{blue}{\underline{0.813 ± 0.024}} \\ \cline{2-11} 
		& \texttt{PRISM}                                     & \multicolumn{1}{l|}{\textcolor{red}{\textbf{200.7 ± 4.8}}}             & \multicolumn{1}{l|}{\textcolor{red}{\textbf{315.1 ± 5.7}}}              & \textcolor{red}{\textbf{0.852 ± 0.005}}              & \multicolumn{1}{l|}{\textcolor{red}{\textbf{819 ± 22}}}             & \multicolumn{1}{l|}{\textcolor{red}{\textbf{1677 ± 46}}}              & \textcolor{red}{\textbf{0.904 ± 0.006}}              & \multicolumn{1}{l|}{\textcolor{red}{\textbf{204.0 ± 10.2}}}             & \multicolumn{1}{l|}{\textcolor{red}{\textbf{364.2 ± 15.0}}}              &\textcolor{red}{\textbf{0.866 ± 0.013}}               \\ \hline \hline 
		\multirow{11}{*}{\textbf{\rotatebox{90}{Crime}}}        & \texttt{MVURE}                                         & \multicolumn{1}{l|}{67.9 ± 1.1}   & \multicolumn{1}{l|}{93.8 ± 1.9}    & 0.591 ± 0.016 & \multicolumn{1}{l|}{100.4 ± 6.6}  & \multicolumn{1}{l|}{129.2 ± 7.3}   & 0.461 ± 0.062 & \multicolumn{1}{l|}{130.3 ± 1.7}  & \multicolumn{1}{l|}{201.7 ± 3.2}   & 0.594 ± 0.013 \\ \cline{2-11} 
		& \texttt{MGFN}                                          & \multicolumn{1}{l|}{70.2 ± 2.3}   & \multicolumn{1}{l|}{89.6 ± 2.5}    & 0.630 ± 0.020 & \multicolumn{1}{l|}{107.4 ± 5.4}  & \multicolumn{1}{l|}{137.9 ± 5.2}   & 0.386 ± 0.047 & \multicolumn{1}{l|}{128.4 ± 3.3}  & \multicolumn{1}{l|}{199.9 ± 4.3}   & 0.601 ± 0.017 \\ \cline{2-11} 
		& \texttt{RDCL}                                     & \multicolumn{1}{l|}{98.7 ± 3.1}   & \multicolumn{1}{l|}{127.9 ± 5.2}   & 0.251 ± 0.026 & \multicolumn{1}{l|}{121.7 ± 4.8}  & \multicolumn{1}{l|}{159.6 ± 6.3}   & 0.179 ± 0.053 & \multicolumn{1}{l|}{156.3 ± 2.1}  & \multicolumn{1}{l|}{242.3 ± 4.6}   & 0.413 ± 0.021 \\ \cline{2-11} 
		& \texttt{HREP}                                          & \multicolumn{1}{l|}{62.8 ± 2.1}   & \multicolumn{1}{l|}{83.1 ± 2.3}    & 0.680 ± 0.014 & \multicolumn{1}{l|}{88.3 ± 6.4}   & \multicolumn{1}{l|}{114.4 ± 5.5}   & 0.578 ± 0.041 & \multicolumn{1}{l|}{124.4 ± 2.3}  & \multicolumn{1}{l|}{196.9 ± 3.9}   & 0.612 ± 0.014 \\ \cline{2-11} 
		
		& \texttt{BiST}                                      & \multicolumn{1}{l|}{64.7 ± 1.1}   & \multicolumn{1}{l|}{83.8 ± 1.4}    & 0.679 ± 0.012 & \multicolumn{1}{l|}{91.3 ± 1.8}   & \multicolumn{1}{l|}{123.2 ± 5.6}   & 0.510 ± 0.004 & \multicolumn{1}{l|}{173.5 ± 4.3}  & \multicolumn{1}{l|}{241.7 ± 6.9}   & 0.415 ± 0.002 \\ \cline{2-11} 
		& \texttt{START}                                      & \multicolumn{1}{l|}{72.6 ± 1.5}   & \multicolumn{1}{l|}{98.9 ± 1.6}    & 0.552 ± 0.147& \multicolumn{1}{l|}{93.0 ± 3.7}   & \multicolumn{1}{l|}{128.9 ± 4.6}   & 0.464 ± 0.039 & \multicolumn{1}{l|}{114.4 ± 3.5}  & \multicolumn{1}{l|}{202.1 ± 8.0}   & 0.591 ± 0.033 \\ \cline{2-11} 
		& \texttt{UrbanMoE}                                      & \multicolumn{1}{l|}{97.1 ± 1.6}   & \multicolumn{1}{l|}{131.0 ± 1.5}    & 0.215 ± 0.017 & \multicolumn{1}{l|}{121.3 ± 3.2}   & \multicolumn{1}{l|}{155.7 ± 3.3}   & 0.218 ± 0.032 & \multicolumn{1}{l|}{138.7 ± 4.2}  & \multicolumn{1}{l|}{252.8 ± 7.3}   & 0.361 ± 0.037 \\ \cline{2-11} 
		& \texttt{HGN}                                      & \multicolumn{1}{l|}{\textcolor{blue}{\underline{55.2 ± 6.5}}}    & \multicolumn{1}{l|}{76.9 ± 15.8}     & 0.729 ± 0.001 & \multicolumn{1}{l|}{93.5 ± 14.1} & \multicolumn{1}{l|}{129.7 ± 17.8}  &0.457 ± 0.001 & \multicolumn{1}{l|}{103.5 ± 6.2}   & \multicolumn{1}{l|}{\textcolor{blue}{\underline{176.4 ± 12.8}}}   &\textcolor{blue}{\underline{0.689 ± 0.001}} \\ \cline{2-11} 
		& \texttt{PGN}                                      & \multicolumn{1}{l|}{56.1 ± 1.1}    & \multicolumn{1}{l|}{78.8 ± 0.2}     & 0.716 ± 0.002 & \multicolumn{1}{l|}{83.1 ± 1.6}  & \multicolumn{1}{l|}{117.4 ± 1.9}  & 0.556 ± 0.015& \multicolumn{1}{l|}{106.6 ± 5.6}  & \multicolumn{1}{l|}{183.6 ± 3.8}   & 0.663 ± 0.013 \\ \cline{2-11} 
		& \texttt{HAF}                                      & \multicolumn{1}{l|}{56.1 ± 1.3}   & \multicolumn{1}{l|}{\textcolor{blue}{\underline{76.1 ± 2.2}}}    & \textcolor{blue}{\underline{0.734 ± 0.015}} & \multicolumn{1}{l|}{\textcolor{blue}{\underline{77.8 ± 3.6}}}   & \multicolumn{1}{l|}{\textcolor{blue}{\underline{107.1 ± 5.4}}}   & \textcolor{blue}{\underline{0.631 ± 0.036}} & \multicolumn{1}{l|}{\textcolor{blue}{\underline{101.5 ± 3.3}}}  & \multicolumn{1}{l|}{178.4 ± 3.6}   & 0.682 ± 0.013\\ \cline{2-11} 
		& \texttt{PRISM}                                     & \multicolumn{1}{l|}{\textcolor{red}{\textbf{53.9 ± 0.3}}}             & \multicolumn{1}{l|}{\textcolor{red}{\textbf{73.2 ± 0.5}}}              &\textcolor{red}{\textbf{0.756 ± 0.004}}               & \multicolumn{1}{l|}{\textcolor{red}{\textbf{74.4 ± 4.1}}}             & \multicolumn{1}{l|}{\textcolor{red}{\textbf{101.4 ± 5.5}}}              &\textcolor{red}{\textbf{0.668 ± 0.035}}               & \multicolumn{1}{l|}{\textcolor{red}{\textbf{96.5 ± 4.4}}}             & \multicolumn{1}{l|}{\textcolor{red}{\textbf{176.0 ± 2.8}}}              &\textcolor{red}{\textbf{0.694 ± 0.010}}               \\ \hline \hline 
		\multirow{11}{*}{\textbf{\rotatebox{90}{Service call}}} & \texttt{MVURE}                                         & \multicolumn{1}{l|}{1428 ± 33}    & \multicolumn{1}{l|}{2180 ± 46}     & 0.367 ± 0.027 & \multicolumn{1}{l|}{190.3 ± 9.8}  & \multicolumn{1}{l|}{266.9 ± 12.1}  & 0.441 ± 0.050 & \multicolumn{1}{l|}{102.1 ± 4.8}  & \multicolumn{1}{l|}{164.7 ± 2.7}   & 0.479 ± 0.017 \\ \cline{2-11} 
		& \texttt{MGFN}                                          & \multicolumn{1}{l|}{1554 ± 81}    & \multicolumn{1}{l|}{2286 ± 115}    & 0.303 ± 0.069 & \multicolumn{1}{l|}{208.2 ± 11.3} & \multicolumn{1}{l|}{293.4 ± 16.6}  & 0.329 ± 0.077 & \multicolumn{1}{l|}{102.8 ± 2.2}  & \multicolumn{1}{l|}{166.3 ± 2.5}   & 0.468 ± 0.021 \\ \cline{2-11} 
		& \texttt{RDCL}                                     & \multicolumn{1}{l|}{1783 ± 21}    & \multicolumn{1}{l|}{2597 ± 38}     & 0.103 ± 0.026 & \multicolumn{1}{l|}{195.7 ± 7.6}  & \multicolumn{1}{l|}{272.1 ± 10.1}  & 0.445 ± 0.041 & \multicolumn{1}{l|}{116.6 ± 2.3}  & \multicolumn{1}{l|}{196.7 ± 3.2}   & 0.256 ± 0.024 \\ \cline{2-11} 
		& \texttt{HREP}                                          & \multicolumn{1}{l|}{1430 ± 29}    & \multicolumn{1}{l|}{2286 ± 34}     & 0.398 ± 0.021 & \multicolumn{1}{l|}{185.7 ± 6.1}  & \multicolumn{1}{l|}{262.2 ± 10.8}  & 0.468 ± 0.022 & \multicolumn{1}{l|}{103.4 ± 3.2}  & \multicolumn{1}{l|}{167.4 ± 4.6}   & 0.461 ± 0.029 \\ \cline{2-11} 
		
		& \texttt{BiST}                                      & \multicolumn{1}{l|}{1458 ± 10}   & \multicolumn{1}{l|}{2137 ± 25}    & 0.393 ± 0.001 & \multicolumn{1}{l|}{193.4 ± 8.7}   & \multicolumn{1}{l|}{269.7 ± 7.7}   & 0.435 ± 0.001 & \multicolumn{1}{l|}{144.1 ± 4.7}  & \multicolumn{1}{l|}{211.8 ± 4.0}   & 0.355 ± 0.018 \\ \cline{2-11} 
		& \texttt{START}                                      & \multicolumn{1}{l|}{1405 ± 34}   & \multicolumn{1}{l|}{2155 ± 40}    & 0.382 ± 0.023 & \multicolumn{1}{l|}{201.6 ± 6.1}   & \multicolumn{1}{l|}{287.5 ± 9.6}   & 0.358 ± 0.044 & \multicolumn{1}{l|}{98.2 ± 3.3}  & \multicolumn{1}{l|}{174.9 ± 2.5}   & 0.412 ± 0.017 \\ \cline{2-11} 
		& \texttt{UrbanMoE}                                      & \multicolumn{1}{l|}{1777 ± 19}   & \multicolumn{1}{l|}{2595 ± 17}    & 0.104 ± 0.011 & \multicolumn{1}{l|}{243.6 ± 6.7}   & \multicolumn{1}{l|}{324.9 ± 7.1}  & 0.180 ± 0.036 & \multicolumn{1}{l|}{111.0 ± 2.5}  & \multicolumn{1}{l|}{196.9 ± 2.4}   &0.255 ± 0.019 \\ \cline{2-11} 
		& \texttt{HGN}                                    & \multicolumn{1}{l|}{\textcolor{blue}{\underline{1273 ± 10}}}    & \multicolumn{1}{l|}{1957 ± 21}     &\textcolor{blue}{\underline{0.499 ± 0.000}} & \multicolumn{1}{l|}{190.6 ± 6.8} & \multicolumn{1}{l|}{270.9 ± 14.4}  & 0.429 ± 0.002 & \multicolumn{1}{l|}{84.9 ± 8.8}   & \multicolumn{1}{l|}{147.0 ± 2.9}   &0.584 ± 0.011 \\ \cline{2-11} 
		& \texttt{PGN}                                     & \multicolumn{1}{l|}{1327 ± 7}    & \multicolumn{1}{l|}{2057 ± 22}     & 0.437 ± 0.012 & \multicolumn{1}{l|}{190.2 ± 5.3}  & \multicolumn{1}{l|}{269.5 ± 7.6}  & 0.436 ± 0.032 & \multicolumn{1}{l|}{87.9 ± 2.0}  & \multicolumn{1}{l|}{153.1 ± 3.1}   & 0.550 ± 0.018 \\ \cline{2-11} 
		& \texttt{HAF}                                      & \multicolumn{1}{l|}{1273 ± 20}    & \multicolumn{1}{l|}{\textcolor{blue}{\underline{1951 ± 27}}}     & 0.493 ± 0.014 & \multicolumn{1}{l|}{\textcolor{blue}{\underline{159.3 ± 13.9}}} & \multicolumn{1}{l|}{\textcolor{blue}{\underline{222.0 ± 18.9}}}  & \textcolor{blue}{\underline{0.613 ± 0.067}} & \multicolumn{1}{l|}{\textcolor{blue}{\underline{81.5 ± 2.5}}}   & \multicolumn{1}{l|}{\textcolor{blue}{\underline{142.1 ± 3.2}}}   & \textcolor{blue}{\underline{0.612 ± 0.018}} \\ \cline{2-11} 
		& \texttt{PRISM}                                     & \multicolumn{1}{l|}{\textcolor{red}{\textbf{1268 ± 1.0}}}             & \multicolumn{1}{l|}{\textcolor{red}{\textbf{1937 ± 16}}}              &\textcolor{red}{\textbf{0.501 ± 0.008}}               & \multicolumn{1}{l|}{\textcolor{red}{\textbf{149.3 ± 3.1}}}             & \multicolumn{1}{l|}{\textcolor{red}{\textbf{212.3 ± 4.2}}}              &\textcolor{red}{
			\textbf{0.650 ± 0.014}}               & \multicolumn{1}{l|}{\textcolor{red}{\textbf{78.9 ± 3.1}}}             & \multicolumn{1}{l|}{\textcolor{red}{\textbf{134.6 ± 10.2}}}              &\textcolor{red}{\textbf{0.651 ± 0.052}}               \\ \bottomrule[1.5pt]
	\end{tabular}
	\label{tab:main}
	\vspace{-10pt}
\end{table*}
\section{Experimental Study}

\subsection{Settings}

\noindent
\textbf{Datasets.} Following the study \texttt{HAFusion}~\cite{DBLP:conf/icde/Sun0CFKT24}, we evaluate \texttt{PRISM} on three real-world datasets, including New York City (\textbf{NYC})\footnote{\url{https://opendata.cityofnewyork.us/} (Last accessed: October 24, 2026).}, Chicago (\textbf{Chi})\footnote{\url{https://data.cityofchicago.org/} (Last accessed: October 24, 2026).}, and San Francisco (\textbf{SF})\footnote{\url{https://datasf.org/opendata/} (Last accessed: October 24, 2026).}.
%
For each city, we construct region-level features from multiple urban data sources:
\textbf{Spatial regions. }Each city is partitioned into regions that serve as basic analytical units.
\textbf{Mobility data. }Taxi trajectories are aggregated into region-level origin--destination (OD) matrices,
capturing directed inter-region human flows.
\textbf{POIs. }POI data is collected from OpenStreetMap\footnote{\url{https://www.openstreetmap.org/} (Last accessed: October 24, 2026).}. Region representations are derived from aggregated semantic category counts.
\textbf{Land use. }Zoning data describes the distribution of land-use categories within each region
(11 for NYC, 12 for Chi, and 13 for SF).
\textbf{Urban activity. }Crime reports, Foursquare check-ins\footnote{\url{https://sites.google.com/site/yangdingqi/home/foursquare-dataset} (Last accessed: October 24, 2026).},
and service call records are aggregated at the region level and used as downstream prediction targets~\cite{DBLP:conf/ijcai/0004LLH20,DBLP:conf/aaai/ZhouHCS023,DBLP:conf/ijcai/WuYFPZZ0W22,DBLP:conf/icde/Sun0CFKT24}.
\noindent
\textbf{Downstream Applications.}
To evaluate the effectiveness of the learned urban region representations, we consider three representative applications: \emph{check-in prediction}, \emph{crime forecasting}, and \emph{service call prediction}. These tasks capture diverse aspects of urban dynamics, including human activity intensity, public safety patterns, and infrastructure demand.
We formulate all tasks as regression problems, where the objective is to predict the aggregated event counts within each region using the learned embeddings as input features. Following existing studies~\cite{DBLP:conf/ijcai/0004LLH20,DBLP:conf/aaai/ZhouHCS023,DBLP:conf/ijcai/WuYFPZZ0W22,DBLP:conf/icde/Sun0CFKT24}, we employ Lasso regression~\cite{DBLP:conf/aisd/KumarSP23} due to its robustness to high-dimensional features and its ability to mitigate overfitting through sparsity-inducing regularization.

\noindent
\textbf{Baselines. }
We compare \texttt{PRISM} with representative state-of-the-art baselines from three categories: urban region representation learning, multi-task optimization, and spatio-temporal prediction. The urban region representation learning baselines include \texttt{MVURE}~\cite{DBLP:conf/ijcai/0004LLH20}, \texttt{MGFN}~\cite{DBLP:conf/ijcai/WuYFPZZ0W22}, \texttt{RegionDCL} (\texttt{RDCL})~\cite{DBLP:conf/kdd/LiHCW023}, \texttt{HREP}~\cite{DBLP:conf/aaai/ZhouHCS023}, \texttt{HAFusion} (\texttt{HAF})~\cite{DBLP:conf/icde/Sun0CFKT24}, and \texttt{UrbanMoE}~\cite{DBLP:conf/www/LiuLZMJLZK26}. For multi-task optimization, we consider \texttt{HAFus\\ion-GN} (i.e., HGN) and \texttt{PRISM-GN} (i.e., PGN), both implemented using GradNorm~\cite{DBLP:conf/icml/ChenBLR18}. 
The spatio-temporal prediction baselines include \texttt{START}~\cite{DBLP:journals/tcss/ButtLAS25} and \texttt{BiST}~\cite{DBLP:journals/pvldb/MaWWZWW25}.

\noindent
\textbf{Evaluation Metrics and Implementation Details.} We evaluate all tasks using MAE, RMSE, and $R^2$. All model training experiments are conducted on Nvidia Quadro RTX 8000 GPUs with 48 GB memory, and the code environment is based on the PyTorch framework using Python 3.8.18.

\begin{table*}[t]
	\centering
	\small
	\caption{Ablation study for \texttt{PRISM}. For each dataset, the best results are in bold/red, and the second-best results are underlined/in blue.}
	\begin{tabular}{l|l|lll|lll|lll}
		\toprule[1.5pt]
		\multirow{2}{*}{\textbf{}}             & \multicolumn{1}{c|}{\multirow{2}{*}{\textbf{Method}}} & \multicolumn{3}{c|}{\textbf{NYC}}                                            & \multicolumn{3}{c|}{\textbf{Chi}}                                                  & \multicolumn{3}{c}{\textbf{SF}}                                            \\ \cline{3-11} 
		& \multicolumn{1}{c|}{}                                  & \multicolumn{1}{c|}{\textbf{MAE} ($\downarrow$)} & \multicolumn{1}{c|}{\textbf{RMSE} ($\downarrow$)} & \multicolumn{1}{c|}{\textbf{$R^2$} ($\uparrow$)}  & \multicolumn{1}{c|}{\textbf{MAE} ($\downarrow$)} & \multicolumn{1}{c|}{\textbf{RMSE}$\downarrow$} &\multicolumn{1}{c|}{ \textbf{$R^2$} ($\uparrow$)}   & \multicolumn{1}{c|}{\textbf{MAE} ($\downarrow$)} & \multicolumn{1}{c|}{\textbf{RMSE} ($\downarrow$)} & \multicolumn{1}{c}{\textbf{$R^2$} ($\uparrow$)}   \\ \hline \hline
		\multirow{4}{*}{\textbf{\rotatebox{90}{Check-in}}}     
		& \texttt{w/o RAR}                                      & \multicolumn{1}{l|}{203.5 ± 1.6}    & \multicolumn{1}{l|}{331.1 ± 0.1}     &0.836 ± 0.000  & \multicolumn{1}{l|}{1012 ± 21.8} & \multicolumn{1}{l|}{2096 ± 52.0}  &0.850 ± 0.007  & \multicolumn{1}{l|}{\textcolor{blue}{\underline{231.6 ± 31.4}}}   & \multicolumn{1}{l|}{435.1 ± 23.1}   &0.809 ± 0.020  \\ \cline{2-11} 
		& \texttt{w/o UM}                                      & \multicolumn{1}{l|}{210.3 ± 8.7}    & \multicolumn{1}{l|}{336.7 ± 12.7}     &0.831 ± 0.013  & \multicolumn{1}{l|}{950 ± 21.4} & \multicolumn{1}{l|}{2057 ± 19.3}  &0.856 ± 0.003  & \multicolumn{1}{l|}{232.1 ± 12.5}   & \multicolumn{1}{l|}{444.6 ± 44.9}   &0.800 ± 0.040  \\ \cline{2-11} 
		& \texttt{w/o PID}                                      & \multicolumn{1}{l|}{\textcolor{blue}{\underline{202.9 ± 2.6}}}    & \multicolumn{1}{l|}{\textcolor{blue}{\underline{315.5 ± 4.1}}}     &\textcolor{blue}{\underline{0.852 ± 0.014}}  & \multicolumn{1}{l|}{\textcolor{blue}{\underline{975 ± 40.0}}} & \multicolumn{1}{l|}{\textcolor{blue}{\underline{1973 ± 45}}}  &\textcolor{blue}{\underline{0.867 ± 0.006}}  & \multicolumn{1}{l|}{237.3 ± 4.3}   & \multicolumn{1}{l|}{\textcolor{blue}{\underline{422.3 ± 18.4}}}   &\textcolor{blue}{\underline{0.820 ± 0.016}}  \\ \cline{2-11}  
		& \texttt{PRISM}                                     & \multicolumn{1}{l|}{\textcolor{red}{\textbf{200.7 ± 4.8}}}             & \multicolumn{1}{l|}{\textcolor{red}{\textbf{315.1 ± 5.7}}}              & \textcolor{red}{\textbf{0.852 ± 0.005}}              & \multicolumn{1}{l|}{\textcolor{red}{\textbf{819 ± 22.0}}}             & \multicolumn{1}{l|}{\textcolor{red}{\textbf{1677 ± 46}}}              & \textcolor{red}{\textbf{0.904 ± 0.006}}              & \multicolumn{1}{l|}{\textcolor{red}{\textbf{204.0 ± 10.2}}}             & \multicolumn{1}{l|}{\textcolor{red}{\textbf{364.2 ± 15.0}}}              &\textcolor{red}{\textbf{0.866 ± 0.013}}               \\ \hline \hline 
		
		\multirow{4}{*}{\textbf{\rotatebox{90}{Crime}}}        
		& \texttt{w/o RAR}                                      & \multicolumn{1}{l|}{ 56.9 ± 3.1}    & \multicolumn{1}{l|}{76.5 ± 3.1}     &0.733 ± 0.022  & \multicolumn{1}{l|}{ 78.5 ± 3.8} & \multicolumn{1}{l|}{\textcolor{blue}{\underline{105.4 ± 3.7}}}  &\textcolor{blue}{\underline{0.642 ± 0.025}}  & \multicolumn{1}{l|}{103.2 ± 5.9}   & \multicolumn{1}{l|}{179.3 ± 1.3}   &0.679 ± 0.005  \\ \cline{2-11} 
		& \texttt{w/o UM}                                      & \multicolumn{1}{l|}{ 57.6 ± 3.2}    & \multicolumn{1}{l|}{ 77.3 ± 5.0}     &0.726 ± 0.035
		& \multicolumn{1}{l|}{83.3 ± 2.0} & \multicolumn{1}{l|}{114.3 ± 2.4}  &0.579 ± 0.018  & \multicolumn{1}{l|}{102.7 ± 1.1}   & \multicolumn{1}{l|}{176.2 ± 6.4}   &0.690 ± 0.023  \\ \cline{2-11} 
		& \texttt{w/o PID}                                      & \multicolumn{1}{l|}{\textcolor{blue}{\underline{56.3 ± 1.3}}}    & \multicolumn{1}{l|}{\textcolor{blue}{\underline{75.0 ± 0.7}}}     &\textcolor{blue}{\underline{0.743 ± 0.004}}  & \multicolumn{1}{l|}{\textcolor{blue}{\underline{84.1 ± 6.4}}} & \multicolumn{1}{l|}{113.8 ± 2.8}  &0.583 ± 0.021  & \multicolumn{1}{l|}{\textcolor{blue}{\underline{102.0 ± 2.1}}}   & \multicolumn{1}{l|}{\textcolor{blue}{\underline{171.9 ± 11.4}}}   &\textcolor{blue}{\underline{0.705 ± 0.039}}  \\ \cline{2-11} 
		& \texttt{PRISM}                                     & \multicolumn{1}{l|}{\textcolor{red}{\textbf{53.9 ± 0.3}}}             & \multicolumn{1}{l|}{\textcolor{red}{\textbf{73.2 ± 0.5}}}              &\textcolor{red}{\textbf{0.756 ± 0.004}}               & \multicolumn{1}{l|}{\textcolor{red}{\textbf{74.4 ± 4.1}}}             & \multicolumn{1}{l|}{\textcolor{red}{\textbf{101.4 ± 5.5}}}              &\textcolor{red}{\textbf{0.668 ± 0.035}}               & \multicolumn{1}{l|}{\textcolor{red}{\textbf{96.5 ± 4.4}}}             & \multicolumn{1}{l|}{\textcolor{red}{\textbf{176.0 ± 2.8}}}              &\textcolor{red}{\textbf{0.694 ± 0.010}}               \\ \hline \hline 
		
		\multirow{4}{*}{\textbf{\rotatebox{90}{Service call}}} 
		& \texttt{w/o RAR}                                      & \multicolumn{1}{l|}{1311 ± 40.6}    & \multicolumn{1}{l|}{1974 ± 15.9}     &0.482 ± 0.009  & \multicolumn{1}{l|}{164.8 ± 14.5} & \multicolumn{1}{l|}{236.0 ± 11.7}  &0.567 ± 0.042  & \multicolumn{1}{l|}{81.7 ± 2.4}   & \multicolumn{1}{l|}{139.1 ± 5.8}   &0.628 ± 0.031  \\ \cline{2-11} 
		& \texttt{w/o UM}                                      & \multicolumn{1}{l|}{1316 ± 15.9}    & \multicolumn{1}{l|}{1967 ± 34.1}     &0.486 ± 0.018  & \multicolumn{1}{l|}{169.4 ± 4.3} & \multicolumn{1}{l|}{233.4 ± 16.8}  &0.576 ± 0.061  & \multicolumn{1}{l|}{82.1 ± 1.1}   & \multicolumn{1}{l|}{141.0 ± 2.1}   &0.618 ± 0.011  \\ \cline{2-11} 
		& \texttt{w/o PID}                                      & \multicolumn{1}{l|}{\textcolor{blue}{\underline{1296 ± 7.8}}}    & \multicolumn{1}{l|}{\textcolor{blue}{\underline{1962 ± 31.1}}}     &\textcolor{blue}{\underline{0.489 ± 0.016}}  & \multicolumn{1}{l|}{\textcolor{blue}{\underline{162.4 ± 9.0}}} & \multicolumn{1}{l|}{\textcolor{blue}{\underline{225.8 ± 19.7}}}  &\textcolor{blue}{\underline{0.603 ± 0.069}}  & \multicolumn{1}{l|}{\textcolor{blue}{\underline{80.6 ± 3.2}}}   & \multicolumn{1}{l|}{\textcolor{blue}{\underline{137.3 ± 9.7}}}   &\textcolor{blue}{\underline{0.637 ± 0.051}}  \\ \cline{2-11} 
		& \texttt{PRISM}                                     & \multicolumn{1}{l|}{\textcolor{red}{\textbf{1268 ± 1.0}}}             & \multicolumn{1}{l|}{\textcolor{red}{\textbf{1937 ± 16}}}              &\textcolor{red}{\textbf{0.501 ± 0.008}}               & \multicolumn{1}{l|}{\textcolor{red}{\textbf{149.3 ± 3.1}}}             & \multicolumn{1}{l|}{\textcolor{red}{\textbf{212.3 ± 4.2}}}              &\textcolor{red}{
			\textbf{0.650 ± 0.014}}               & \multicolumn{1}{l|}{\textcolor{red}{\textbf{78.9 ± 3.1}}}             & \multicolumn{1}{l|}{\textcolor{red}{\textbf{134.6 ± 10.2}}}              &\textcolor{red}{\textbf{0.651 ± 0.052}}      \\ \toprule[1.5pt]
	\end{tabular}
	\label{tab:abl}
	\vspace{-8pt}
\end{table*}

\subsection{Overall Performance}
\noindent
\textbf{Check-in Prediction.}
We begin by evaluating all methods on the check-in prediction task, which measures a model’s ability to capture latent urban functionality and mobility dynamics. As can be seen in Table~\ref{tab:main}, \texttt{PRISM} achieves the best performance across all three cities, consistently outperforming the strongest baseline, \texttt{HAF}, on every evaluation metric.
In NYC, \texttt{PRISM} reduces the MAE from 202.8 to 200.7 and improves the $R^2$ from 0.844 to 0.852. Similar improvements are observed in both Chi and SF, where \texttt{PRISM} achieves the lowest prediction errors and the highest explanatory power.
These results indicate that \texttt{PRISM} more effectively models cross-region interactions and heterogeneous urban signals, which are essential for capturing complex check-in behaviors. 
Unlike existing baselines that focus on urban region representation learning, multi-task objective optimization, or spatio-temporal dependency modeling, \texttt{PRISM} jointly captures heterogeneous urban semantics, cross-region relational structures, and task-level interactions, thereby producing more discriminative and transferable region embeddings.

\noindent
\textbf{Crime Forecasting.}
Next, crime forecasting requires modeling subtle spatial dependencies and socio-environmental factors. As shown in Table~\ref{tab:main}, \texttt{PRISM} again delivers the strongest results across all datasets, achieving the lowest MAE and RMSE values while obtaining the highest $R^2$ values.
Relative to \texttt{HAF}, \texttt{PRISM} improves $R^2$ from 0.734 to 0.756 in NYC and from 0.631 to 0.668 in Chi, indicating enhanced predictive capability in explaining crime variability. The moderate performance gains are expected, given the inherently stochastic nature of crime data.
The results suggest that \texttt{PRISM} learns region embeddings that preserve fine-grained spatial semantics well, thereby supporting more accurate modeling of localized crime patterns. 
\noindent
\textbf{Service Call Prediction.}
Finally, service call prediction is challenging due to high demand volatility and weak spatial regularity. Here, \texttt{PRISM} achieves the best performance across all cities.
For example, in Chi, \texttt{PRISM} improves $R^2$ from 0.613 to 0.650, representing a meaningful relative gain. Comparable improvements are observed in SF, where \texttt{PRISM} surpasses the second-best method across all metrics.
The larger performance gap observed for this task suggests that \texttt{PRISM} is particularly effective in complex prediction settings, where robust representation learning is critical. 
%

\subsection{Ablation Studies}
Table~\ref{tab:abl} reports the ablation results of \texttt{PRISM}. 
Overall, the full model achieves strong performance across most settings, but the contribution of each component varies by city and task.

\noindent
\textbf{Effect of Region-Aware Routing (RAR).}
Replacing region-aware routing with simple feature concatenation (\texttt{w/o RAR}) results in the most consistent performance drop across nearly all settings. The degradation is particularly evident for the check-in task, where $R^2$ decreases from $0.852$ to $0.836$ in NYC and from $0.904$ to $0.850$ in Chi. This confirms that static concatenation tends to entangle heterogeneous signals and cannot adapt to region-specific informativeness. In contrast, region-aware routing prioritizes the most informative views for each region dynamically, enabling the learned embeddings to capture functional heterogeneity better. Similar trends observed for crime and service call prediction further suggest that adaptive routing offers a general advantage rather than being task-specific.

\noindent
\textbf{Effect of Uncertainty Weighting (UM).}
Removing uncertainty-based weighting and assigning equal importance to all objectives (\texttt{w/o UM}) yields both lower accuracy and higher variance. For example, for check-in prediction in NYC, the MAE increases to $210.3$ with substantially larger variance compared to the full model. This behavior indicates less stable optimization dynamics. Since urban representation learning relies on multiple supervisory signals derived from different modalities, naïvely combining them with fixed weights can introduce gradient conflicts and bias the model toward particular objectives. Uncertainty weighting mitigates this issue by automatically balancing loss contributions, thereby improving convergence stability and cross-city generalization.

\noindent
\textbf{Effect of PID-MSE (Feedback-Regulated Error Modeling).}
Replacing PID-MSE with standard MSE (\texttt{w/o PID}) also leads to performance degradation in most cases, showing the benefit of feedback-inspired discrepancy modeling. By decomposing prediction errors into proportional, accumulated, and differential components, PID-MSE simultaneously suppresses systematic bias and local oscillations during training. This is particularly beneficial for noisy urban signals, as reflected in the consistent improvements observed for crime and service call prediction. The results suggest that regulating optimization dynamics is as important as improving the fusion strategy itself.

\begin{figure}[t]
	\centering
	\includegraphics[scale=0.18]{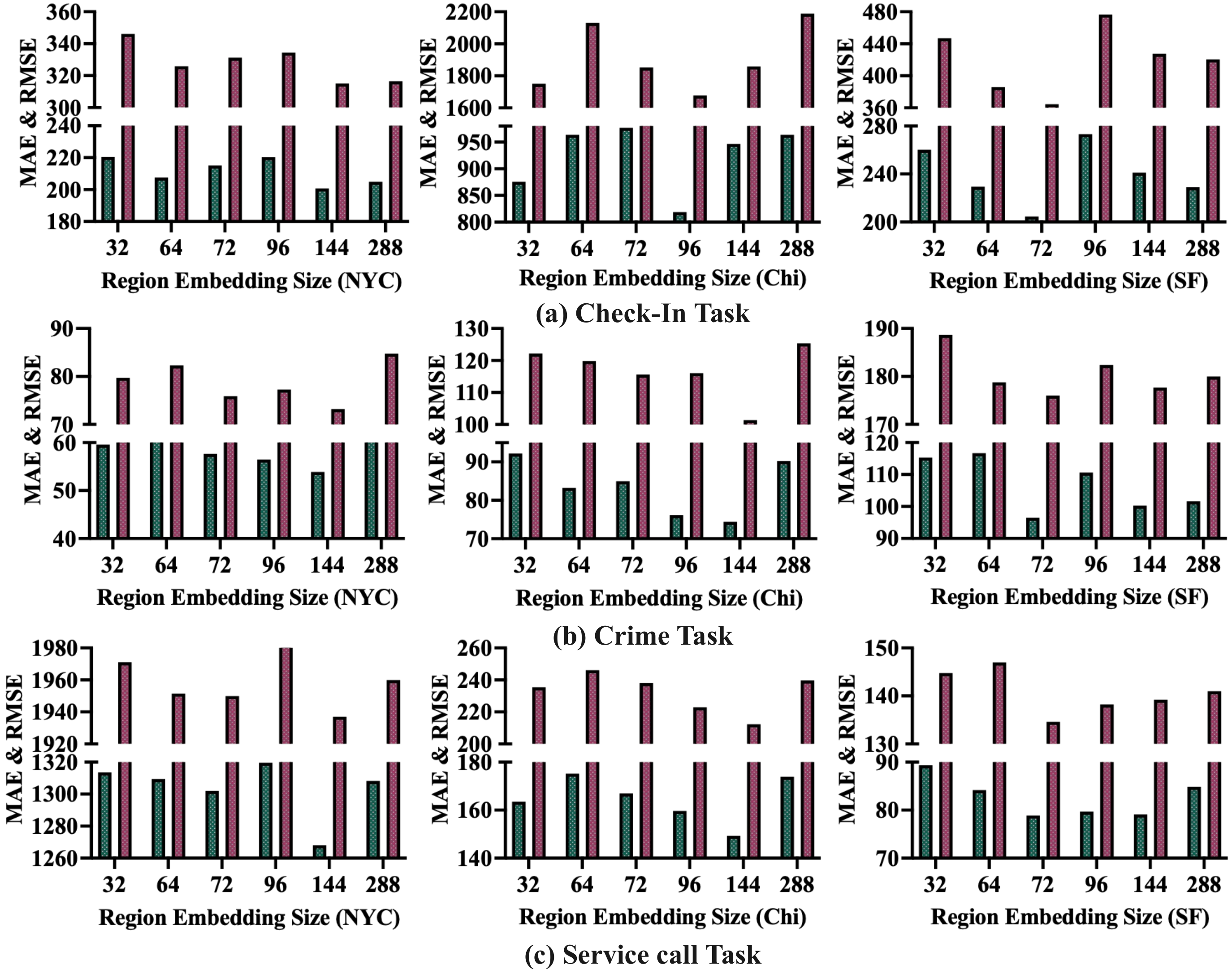}
	\caption{Region Embedding Size Study}
	\label{fig:regionsize}
	\vspace{-15pt}
\end{figure}

\subsection{Parameter Sensitivity Analysis}
\noindent
\textbf{Effects of Region Embedding Size.}
We investigate the impact of region embedding dimensionality on model performance by varying the embedding size from 32 to 288 dimensions. Figure~\ref{fig:regionsize} summarizes the MAE and RMSE results.
Overall, the performance exhibits a non-monotonic trend as the embedding size increases. Very small embeddings (e.g., 32) tend to underperform, suggesting insufficient representational capacity to capture complex spatial semantics and cross-region interactions. As the dimensionality increases, the performance improves consistently, indicating that richer embeddings encode heterogeneous urban signals better.
However, further enlarging the embeddings does not always yield gains. When the dimensionality becomes very large (e.g., 288), the performance either saturates or decreases slightly in several settings. This phenomenon likely stems from over-parameterization, which can introduce redundant features, amplify noise, and increase optimization difficulty.
Interestingly, the optimal embedding size remains relatively stable across cities, typically falling within the mid-range (96–144 dimensions). This suggests that urban region representations require sufficient capacity to model functional diversity but also benefit from controlled compactness to preserve generalization.
From a task perspective, service call prediction appears more sensitive to embedding dimensionality than crime forecasting, possibly because service demand is influenced by a broader set of socio-economic factors that require richer representations.

These findings reveal an important trade-off between representational capacity and model robustness. While larger embeddings enhance expressiveness, overly high dimensionality may harm generalization. 

\begin{figure}[t]
	\centering
	\includegraphics[scale=0.18]{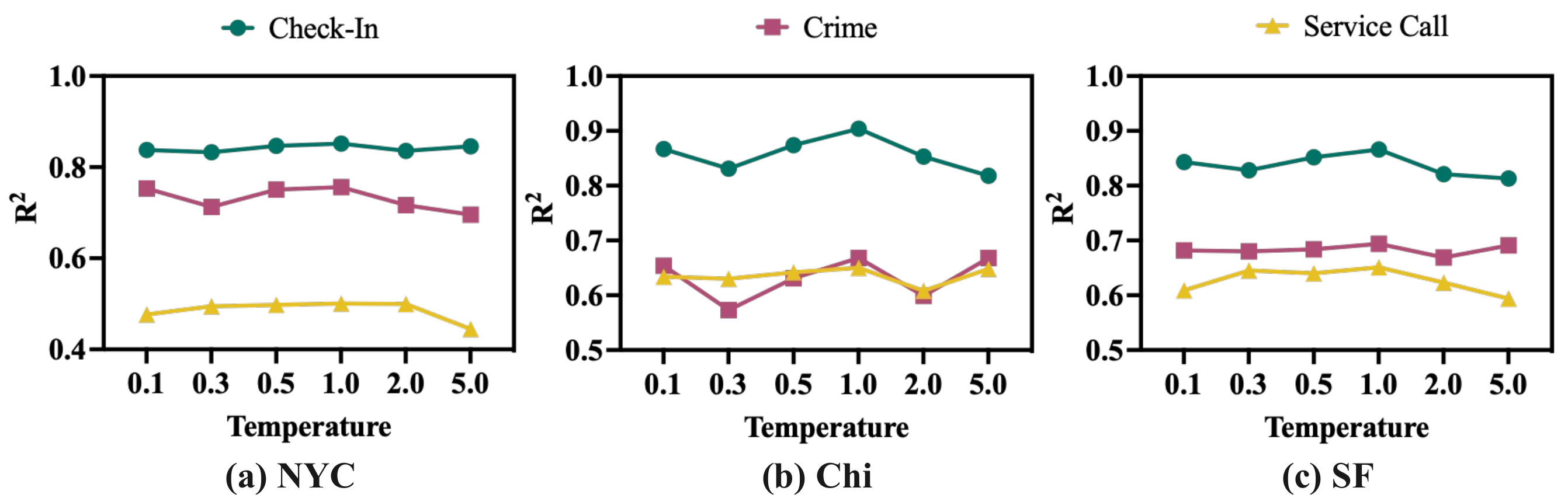}
	\caption{Effect of Router Temperature}
	\label{fig:tem}
	\vspace{-15pt}
\end{figure}

\begin{figure}[t]
	\centering
	\includegraphics[scale=0.2]{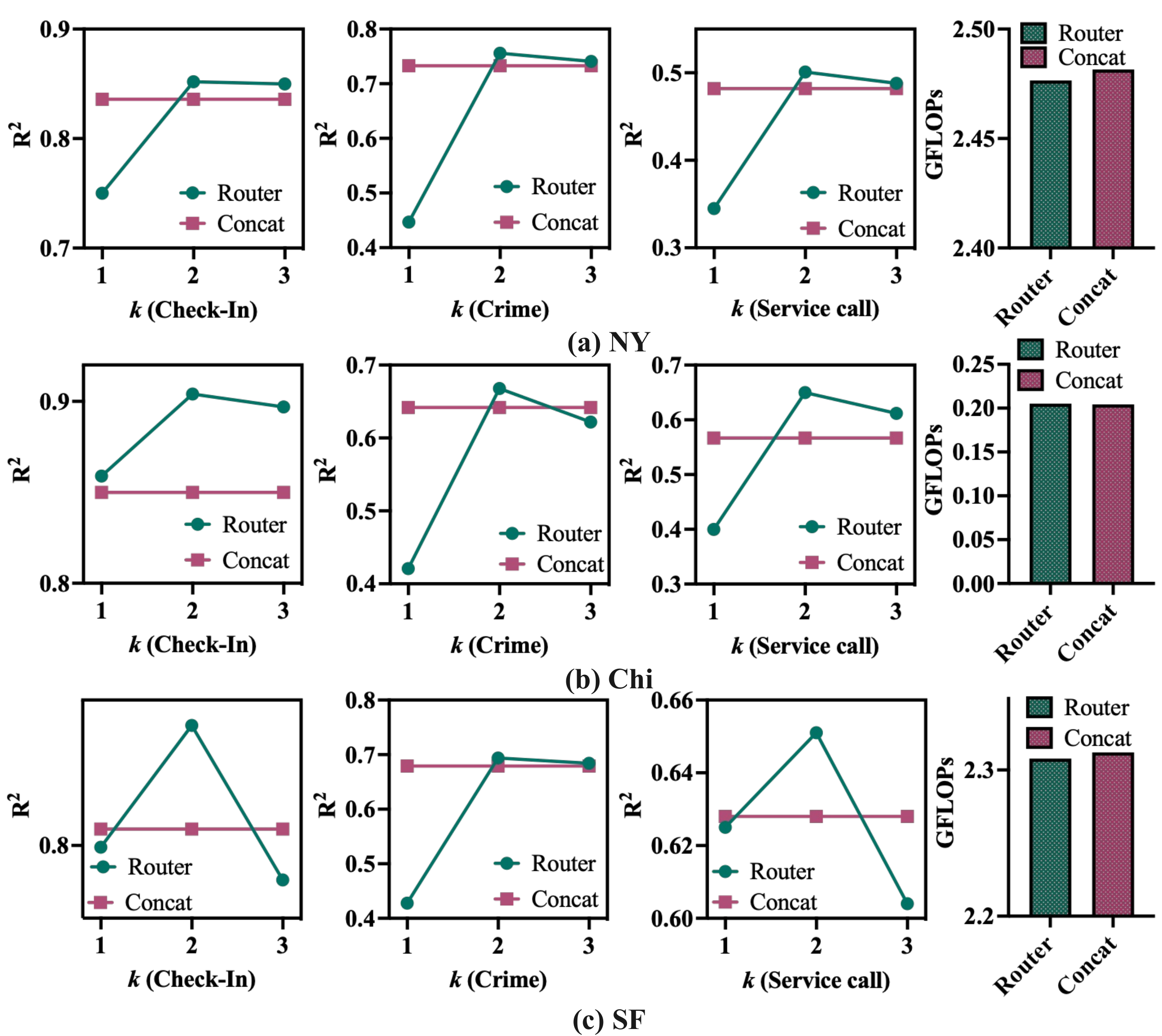}
	\caption{Model Scalability vs. Top-$k$}
	\label{fig:sca-topk}
	\vspace{-10pt}
\end{figure}	

\noindent
\textbf{Effect of Top-$k$ Router Temperature.}
We evaluate the impact of the Top-$k$ router temperature, varying the temperature over $\{0.1, 0.3, 0.5, 1.0, 2.0, 5.0\}$. We report the results using the $R^2$ metric. As shown in Figure~\ref{fig:tem}, prediction performance varies across different temperature settings. Lower temperatures lead to sharp routing, higher temperatures provide smooth routing distributions. We can see that the best results for all three tasks across the three cities occur when the temperature is set to $1.0$.

\subsection{Model Scalability Analysis  }
Figure~\ref{fig:sca-topk} reports on the scalability of \texttt{PRISM}'s Router model under different Top-$k$ settings. We compare the Router model with the Concat baseline (i.e., \texttt{PRISM-Concat})  in terms of prediction performance ($R^2$) and computational cost (GFLOPs).

\noindent
\textbf{Performance vs. Top-$k$.}
We see that, the Router model is sensitive to the choice of $k$. Thus, the prediction performance improves when increasing $k$ from 1 to 2, and then it either stabilizes or decreases slightly when $k=3$. This trend is consistent across multiple settings, suggesting that a moderate number of selected experts provides a good balance between representation capacity and noise control. Specifically, the Router model often achieves its best $R^2$ at $k=2$, outperforming both $k=1$ and $k=3$. When $k=1$, the model may suffer from limited expert diversity, while larger $k$ (e.g., $k=3$) can introduce redundant or less relevant experts, leading to marginal performance degradation.
In contrast, the Concat baseline remains relatively stable across different $k$ values, as it does not explicitly perform dynamic expert selection. However, its performance is generally lower than or comparable to the Router model at the optimal $k$ setting, highlighting the effectiveness of the adaptive routing mechanism.

\noindent
\textbf{Computational Cost.}
The GFLOPs comparison shows that the Router model maintains competitive computational efficiency compared to the Concat baseline. Although increasing $k$ slightly increases computation due to additional expert activation, the overall cost remains comparable to that of Concat. Notably, on the NYC and SF datasets, the Router model even exhibits slightly lower GFLOPs than Concat, indicating that the dynamic routing mechanism can achieve competitive—or even improved—computational efficiency while offering strong predictive performance.

\subsection{Model Efficiency Analysis}
\begin{figure}[t]
	\centering
	\includegraphics[scale=0.25]{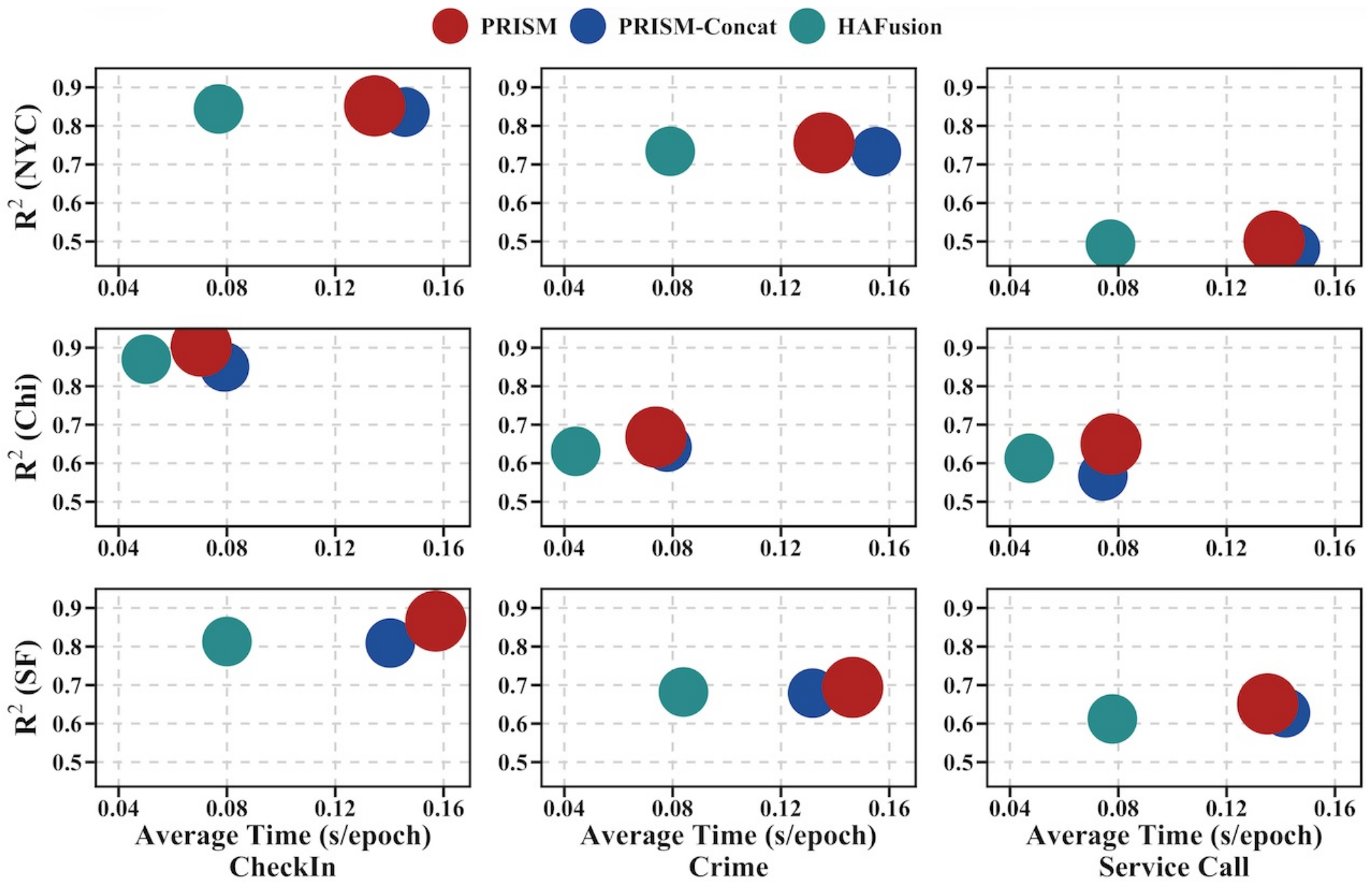}
	\caption{Pareto Analysis of Accuracy --- Efficiency Trade-off }
	\label{fig:eff}
	\vspace{-20pt}
\end{figure}	
Beyond the prediction capabilities, we evaluate the computational efficiency and resource footprint of different fusion strategies. 
Figure~\ref{fig:eff} presents the accuracy--efficiency trade-off across the $3 \times 3$ city--task settings. 
We use $R^2$ to measure prediction performance, report the average training time per epoch (in seconds) to quantify computational cost, and report the peak GPU memory consumption (in GB) to report the footprint. we use marker size to represent the latter. These metrics provide a three-dimensional characterization of computational and memory efficiency.
To quantitatively assess the effectiveness--efficiency trade-off, we define
$
\mathit{Efficiency} = \frac{R^2}{\mathit{Time}}
$, 
where $\mathit{Time}$ denotes the average training time per epoch.

\noindent
\textbf{Pareto Dominance.}
As shown in Figure~\ref{fig:eff}, \texttt{PRISM} consistently lies on or near the Pareto frontier across all nine city–task combinations. In particular, no competing method simultaneously achieves higher prediction accuracy and lower training time. Although \texttt{HAFusion} offers shorter training times, this is accompanied by lower prediction performance. \texttt{PRISM-Concat}, while marginally faster in some scenarios, offers lower accuracy than the routing-based variant.

Importantly, the improvements obtained by \texttt{PRISM} are achieved without noticeable increases in memory consumption, as evidenced by comparable GPU memory usage across methods. This indicates that the adaptive routing mechanism enhances representation quality without introducing significant additional resource overhead.

\noindent
\textbf{Quantitative Efficiency Comparison.}
Across the nine settings, the average efficiency score of \texttt{PRISM} is 6.76, compared to 6.20 for \texttt{PRISM-Concat} and 10.38 for \texttt{HAFusion}. \texttt{HAFusion} achieves a higher numerical efficiency score due to its substantially shorter training time, and the higher score primarily reflects reduced model complexity rather than improved prediction capabilities. In contrast, \texttt{PRISM} consistently achieves better $R^2$ values than does \texttt{PRISM-Concat} while incurring only marginal additional training cost (typically 0.01–0.02 seconds per epoch). From a practical perspective, such minor computational overheads are negligible relative to the observed gains in prediction performance.

\subsection{Model Behavior Analysis}
\label{sec:mba}
To understand the mechanisms underlying the performance improvements offered by \texttt{PRISM}, we analyze its behavior from two perspectives: modality heterogeneity and optimization stability.

\begin{figure}[t]
	\centering
	
	\begin{subfigure}{\linewidth}
		\centering
		\includegraphics[width=1\linewidth]{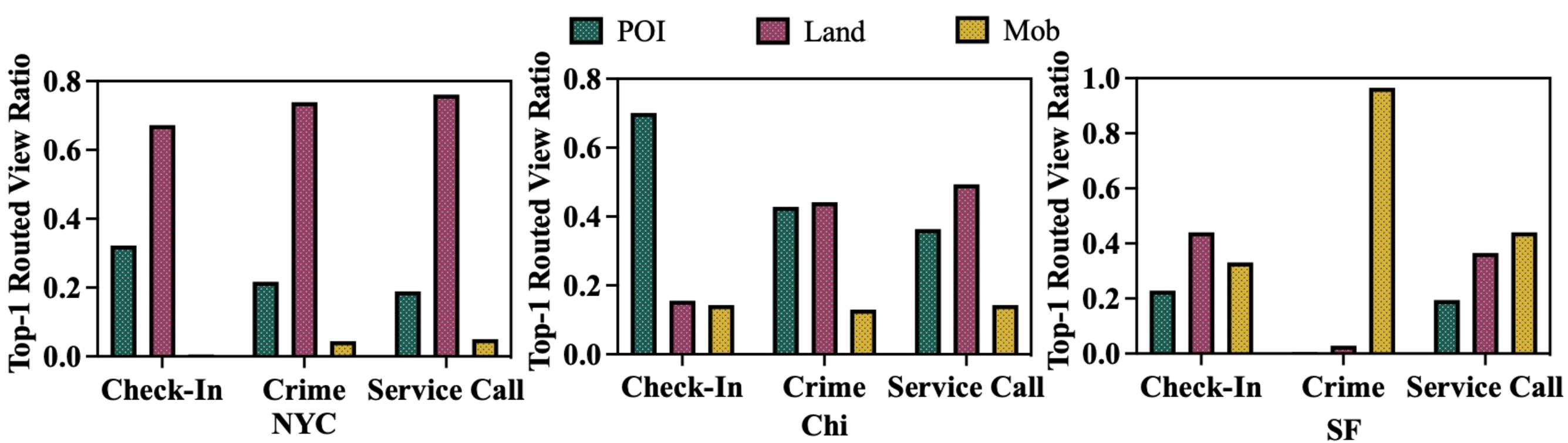}
		\caption{Modality heterogeneity.}
		\label{fig:sub1}
	\end{subfigure}
	
	\vspace{1em}  %
	
	\begin{subfigure}{\linewidth}
		\centering
		\includegraphics[width=1\linewidth]{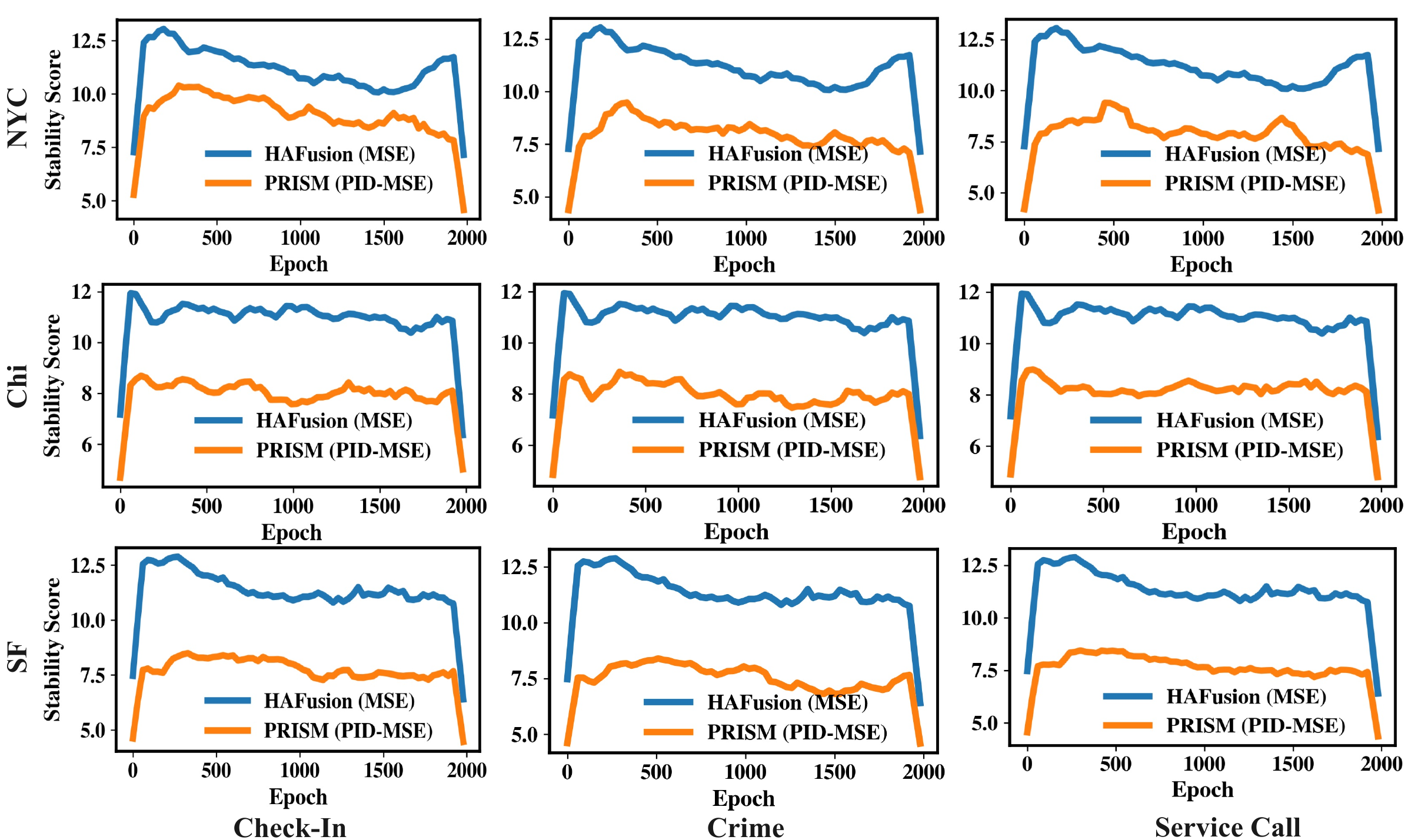}
		\caption{Optimization instability.}
		\label{fig:sub2}
	\end{subfigure}
	
	\caption{Model Behavior Analysis.}
	\label{fig:behavior}
	\vspace{-10pt}
\end{figure}

\noindent
\textbf{Model Heterogeneity.}
Figure~\ref{fig:sub1} reports the Top-1 routed view ratio across different city–task combinations, thus indicating how frequently each modality is selected as the dominant expert. The results show clear task- and city-specific modality preferences. In NYC, the Land modality consistently dominates across all three tasks, suggesting that land-use information plays a primary role in regional prediction. In \textit{Chi}, routing decisions appear more balanced, particularly for Crime and Service Call, where POI and Land exhibit comparable selection frequencies, reflecting stronger cross-modal complementarity. In SF, the Crime task is dominated by the Mobility modality, indicating a strong dependence on human mobility patterns.

These observations indicate substantial modality heterogeneity across city–task settings. The relative importance of modalities varies markedly depending on both the urban context and the prediction objective. Therefore, a fixed fusion strategy is unlikely to perform optimally across all scenarios. In contrast, the adaptive routing mechanism in \texttt{PRISM} adjusts its modality selection dynamically according to task-specific signals, enabling more context-aware representation learning.

\noindent
\textbf{Optimization Instability.}
Figure~\ref{fig:sub2} examines the optimization dynamics by comparing the training stability of \texttt{PRISM} (PID-MSE) and \texttt{HAFusion} (MSE). Across all city–task combinations, \texttt{PRISM} consistently achieves lower instability scores throughout the training. \texttt{HAFusion} exhibits higher and more fluctuating stability values, indicating stronger gradient imbalance and inter-modal conflicts. The performance gap remains stable across epochs, suggesting that the improvement is due to systematic optimization advantages rather than transient effects.
Lower instability scores imply better alignment among modality-specific gradients and reduced scale disparity during joint optimization. By incorporating routing and PID-based regulation, \texttt{PRISM}  mitigates gradient competition across modalities and promotes more coordinated multi-view learning. Consequently, the adaptive routing mechanism not only captures heterogeneous modality relevance but also improves optimization stability, thereby explaining the consistent performance gains observed across city–task settings.

\subsection{Robustness to View Corruption and Incompleteness}
Real-world urban data is inherently prone to view incompleteness and corruption caused by data acquisition limitations and sensor noise. To systematically evaluate the robustness of \texttt{PRISM} and \texttt{HAFusion} under these challenging conditions, we perform experiments in missing-view and noisy-view settings.
\begin{figure}[t]
	\centering
	\includegraphics[scale=0.189]{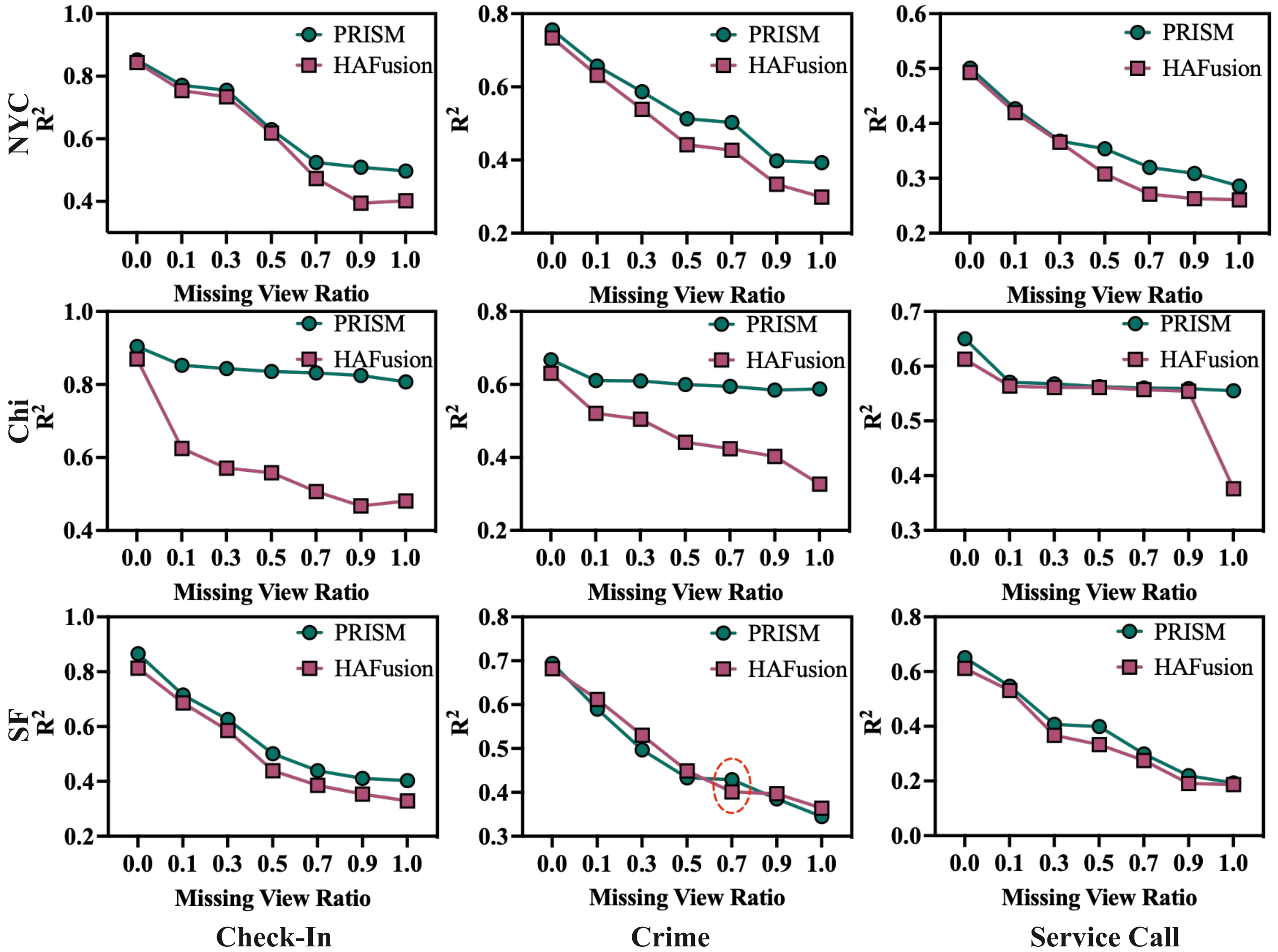}
	\caption{Performance Across Different Missing View Ratio}
	\label{fig:mv}
	\vspace{-15pt}
\end{figure}	

\subsubsection{Missing Views.}
Motivated by the modality heterogeneity analysis in Figure~\ref{fig:sub1}, we investigate the robustness of \texttt{PRISM} in different missing-view scenarios. Specifically, we simulate missing views for the dominant modality in each city: we mask the Land view for NYC, the POI view for Chi, and the Mobility view for SF.
We vary the missing-view ratio in $\{0, 0.1, 0.3, 0.5, 0.7, 0.9, 1.0\}$, where a ratio of 0 indicates no masking and 1.0 corresponds to complete removal of the selected view. Figure 8 reports the prediction performance findings.
As the missing ratio increases, the performance of both methods degrades consistently, confirming the importance of the masked modality. However, \texttt{PRISM} maintains the best performance across nearly all missing ratios and city–task combinations. The performance gap becomes more pronounced under moderate to high missing ratios, indicating that adaptive routing enables the model to better compensate for incomplete information by dynamically reallocating attention to the remaining informative views.
Figure~\ref{fig:mv} also shows that \texttt{HAFusion} slightly outperforms \texttt{PRISM} on SF Crime when the mobility missing ratio is 0.7.
This case reveals a limitation of sparse routing under severe dominant-view incompleteness.
Since SF Crime relies strongly on mobility-related signals, heavily masking the mobility view can make top-$k$ routing decisions less reliable.
By contrast, the dense fusion strategy in \texttt{HAFusion} aggregates all available views and may provide more stable behavior at this specific corruption level.
This result suggests that sparse view selection is generally beneficial, but may require missingness-aware or uncertainty-aware routing when the dominant modality is severely corrupted.
Even when the dominant modality is fully removed (ratio = 1.0), \texttt{PRISM} continues to outperform the baseline in most cases. 

\subsubsection{Noisy Views.}
We evaluate the robustness of \texttt{PRISM} by injecting three types of perturbations—additive Gaussian noise, mixture-of-Gaussians noise, and column-wise shuffling—with corruption ratios varied in $\{0.1, 0.5, 0.9\}$ to simulate progressively severe view degradation. These perturbations collectively capture realistic degradation patterns, including random feature corruption, heterogeneous distribution shifts, and structural information distortion.
Following the missing-view setting, we inject noise into the dominant view of each city: the Land view for NYC, the POI view for Chi, and the Mobility (Mob) view for SF, as these views contribute most significantly to the corresponding downstream tasks in each city.
\begin{figure}[t]
	\centering
	\includegraphics[scale=0.18]{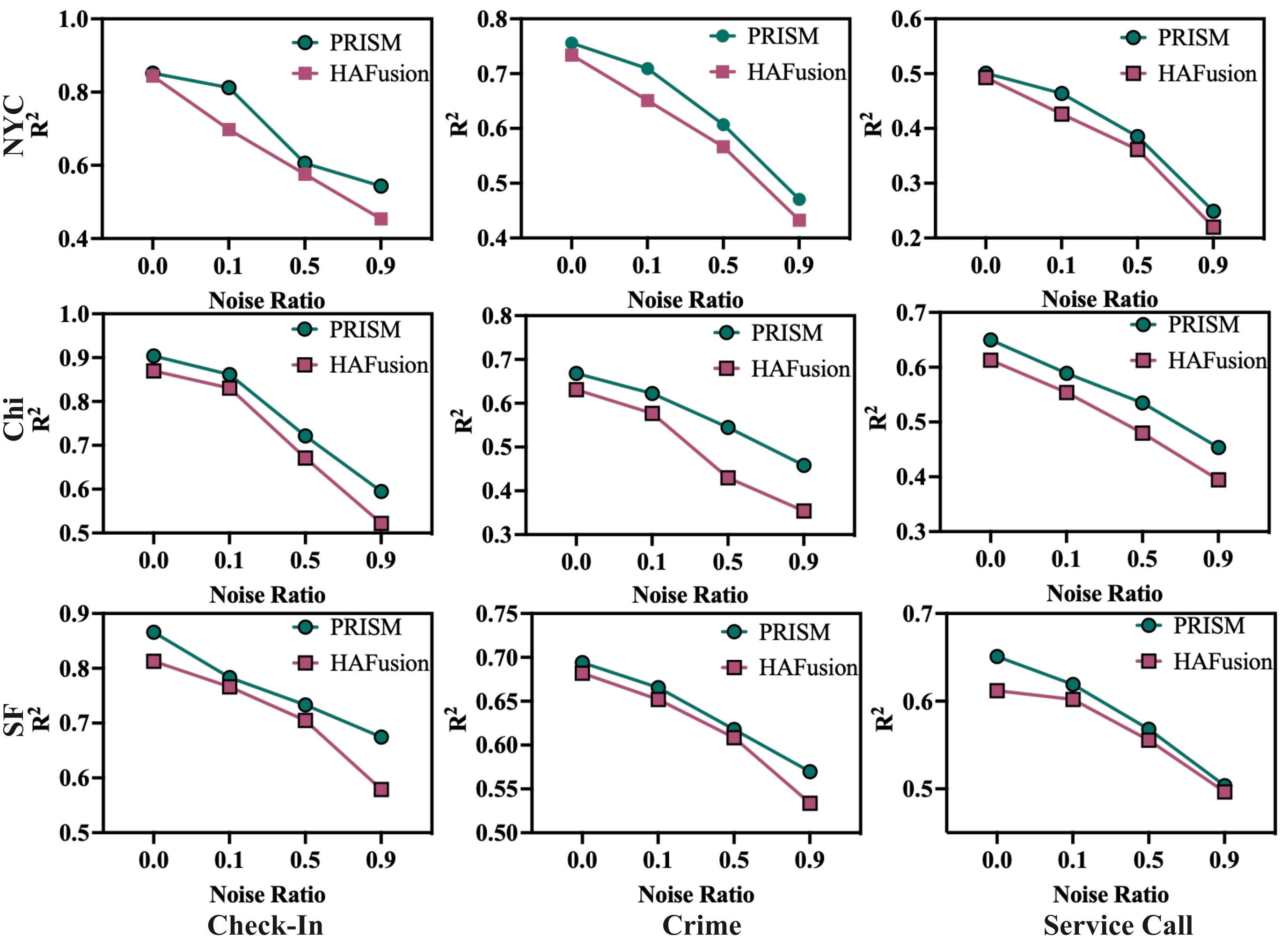}
	\caption{Robustness Analysis --- Gaussian Perturbation}
	\label{fig:gaussian}
	\vspace{-10pt}
\end{figure}	

\noindent
\textbf{Gaussian Noise.} Figure~\ref{fig:gaussian} presents the performance under additive Gaussian perturbations across three cities and three tasks. In NYC, performance declines steadily as the corruption ratio increases, with the most pronounced degradation observed in the Crime task; nevertheless, \texttt{PRISM} consistently maintains higher $R^2$ scores than \texttt{HAFusion}, and the performance gap widens under severe noise ($0.9$). In Chicago, both methods exhibit sharper drops compared to NYC, particularly for Crime and Service Call prediction; however, \texttt{PRISM} demonstrates slower performance decay, indicating improved robustness under moderate-to-high corruption levels. For San Francisco, the degradation trend is comparatively smoother across all three tasks, and \texttt{PRISM} preserves a stable margin over the baseline even at the highest noise ratio, suggesting stronger resistance to stochastic perturbations across diverse urban prediction scenarios.
\begin{figure}[t]
	\centering
	\includegraphics[scale=0.18]{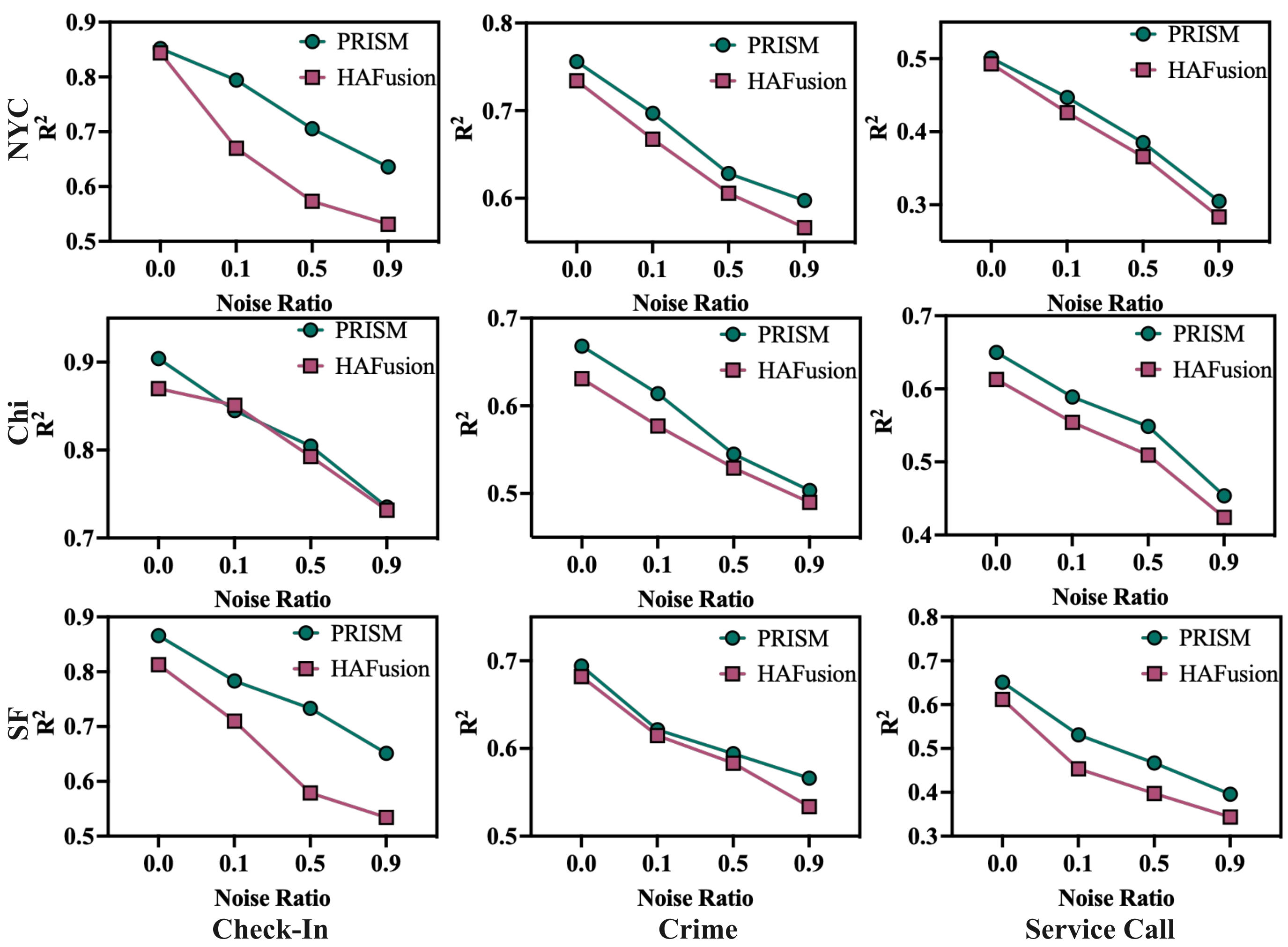}
	\caption{Robustness Analysis --- Multi-Gaussian Perturbation}
	\label{fig:multigaussian}
		\vspace{-15pt}
\end{figure}

\begin{figure}[t]
	\centering
	\includegraphics[scale=0.18]{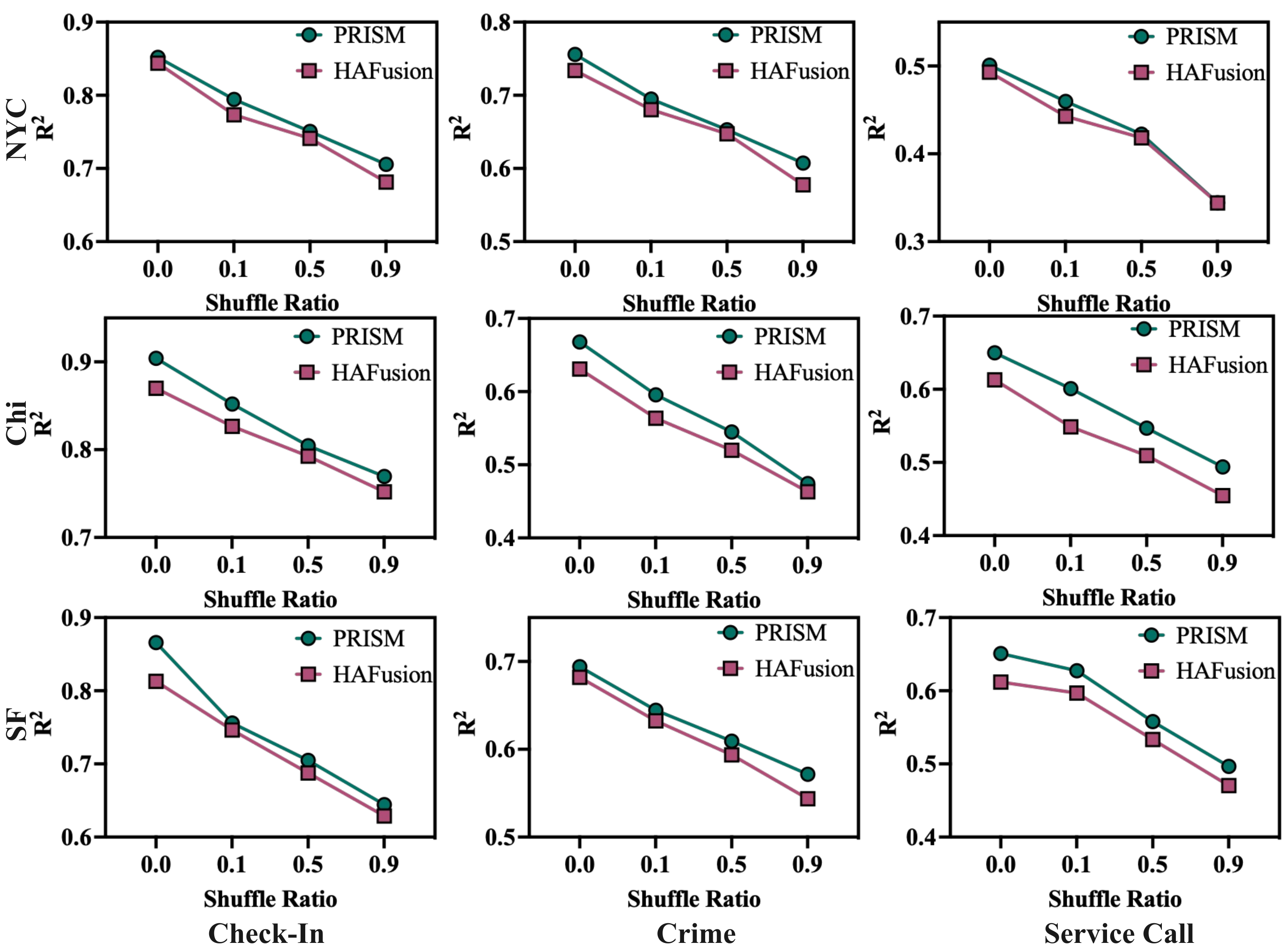}
	\caption{Robustness Analysis --- Column-Wise Shuffle.}
	\label{fig:col}
		\vspace{-10pt}
\end{figure}	

\noindent
\textbf{Multi-Gaussian Noise. }Figure~\ref{fig:multigaussian} presents the results for mixture-of-Gaussians perturbations. In NYC, performance degrades monotonically with increasing corruption, and \texttt{PRISM} achieves more competitive performance than the strongest baseline \texttt{HAFusion} across all three tasks. In Chi, both methods decline as noise intensifies; notably, for the Check-in task at ratio $0.1$, \texttt{HAFusion} slightly surpasses \texttt{PRISM}, whereas \texttt{PRISM} is the best at higher corruption levels. For Crime and Service Call prediction in Chi, \texttt{PRISM} remains consistently better. In SF, degradation is relatively smooth, and \texttt{PRISM} preserves higher $R^2$ scores across all tasks and noise ratios.

\noindent
\textbf{Column-Wise Shuffle.} Figure~\ref{fig:col} presents the results for column-wise shuffling, which permutes feature dimensions to disrupt intra-view structural dependencies. In NYC, both methods exhibit nearly identical degradation patterns across Check-in, Crime, and Service Call prediction. Performance decreases consistently as the shuffle ratio increases, while the gap between \texttt{PRISM} and \texttt{HAFusion} remains marginal across all corruption levels.
In contrast, for Chi and SF, \texttt{PRISM} consistently achieves higher $R^2$ scores across all three tasks and shuffle ratios. The performance gap becomes more pronounced under moderate-to-high shuffle ratios, particularly for Crime and Service Call prediction, indicating improved stability under severe structural perturbation.

\section{Conclusion }
We present \texttt{PRISM}, a self-adjusting framework for multi-view urban region representation learning. By integrating region-wise sparse routing with PID-inspired discrepancy regulation, \texttt{PRISM} enables adaptive modality allocation and stabilizes multi-objective optimization. Extensive experiments on three large-scale urban datasets across three tasks exhibit consistent performance improvements over state-of-the-art proposals while maintaining comparable model complexity.
Our study highlights the importance of adaptive fusion and feedback-aware optimization in heterogeneous urban representation learning, particularly for achieving stable training and preserving model robustness. 
Overall, the results show that adaptive mechanisms are essential for task-specific urban region representation learning. Future work will focus on cross-city generalization and
adaptive fusion in dynamic urban environments.

\section*{Acknowledgements}
This work was supported by the National Natural Science Foundation of China (Nos. 62472174 and 62402082), the ECNU Multifunctional Platform for Innovation (No. 001), and the Chongqing Municipal Education Commission (No. KJQN202400637). 

\balance

\bibliographystyle{ACM-Reference-Format}
\bibliography{sample}

\end{document}